\documentclass[10pt,twocolumn,letterpaper]{article}

\usepackage[pagenumbers]{wacv} % To force page numbers, e.g. for an arXiv version

\usepackage{graphicx}
\usepackage{amsmath}
\usepackage{amssymb}
\usepackage{booktabs}
\usepackage[pagebackref,breaklinks,colorlinks]{hyperref}

\usepackage[capitalize]{cleveref}
\crefname{section}{Sec.}{Secs.}
\Crefname{section}{Section}{Sections}
\Crefname{table}{Table}{Tables}
\crefname{table}{Tab.}{Tabs.}

\def\wacvPaperID{554} % *** Enter the WACV Paper ID here
\def\confName{WACV}
\def\confYear{2024}

\usepackage{mathtools}
\usepackage{booktabs}
\usepackage{multirow}
\usepackage[x11names]{xcolor}

\usepackage[usestackEOL]{stackengine}
\edef\tmp{\the\baselineskip}
\usepackage{rotating}

\usepackage{colortbl}
\newcommand{\myparagraph}[1]{\noindent\textbf{#1}}

\DeclareMathOperator{\similarity}{sim}

\DeclareMathOperator*{\argmax}{arg\;max}
\newcommand{\z}{\mathbf{z}}

\newcommand{\x}{\mathbf{x}}
\newcommand{\f}{\mathbf{f}}

\newcommand{\SupCon}{\text{SupCon}\xspace}
\newcommand{\CE}{\text{CE}\xspace}
\newcommand{\SupTT}{\text{PSupCon}\xspace}
\newcommand{\ours}{\text{OPSupCon}\xspace}
\newcommand{\oursreal}{\text{OPSupCon-R}\xspace}
\newcommand{\oursfake}{\text{OPSupCon-P}\xspace}
\newcommand{\FPR}{\text{FPR}\xspace}
\newcommand{\AUROC}{\text{AUROC}\xspace}
\newcommand{\AUPR}{\text{AUPR}\xspace}
\newcommand{\InDistribution}{\text{ID}\xspace}
\newcommand{\ID}{\text{ID}\xspace}
\newcommand{\OOD}{\text{OOD}\xspace}

\renewcommand{\th}{\boldsymbol{\theta}}

\newcommand{\fprcolor}{Honeydew1} 
\newcommand{\auroccolor}{Honeydew2}
\newcommand{\auprcolor}{Honeydew3}
\newcommand{\metricsize}{0.4cm}

\begin{document}

\title{OOD Aware Supervised Contrastive Learning}

\author{Soroush Seifi\\
{\tt\small soroush.seifi@external.toyota-europe.com}
\and 
Daniel Olmeda Reino\\
{\tt\small daniel.olmeda.reino@toyota-europe.com}
\and 
Nikolay Chumerin\\
{\tt\small nikolay.chumerin@toyota-europe.com}
\and
Rahaf Aljundi\\
{\tt\small rahaf.al.jundi@toyota-europe.com}\\
\and
Toyota Motor Europe
}

\maketitle

%%%%%%%%% ABSTRACT
\begin{abstract}
\vspace{-0.3cm}
Out-of-Distribution~(\OOD) detection is a crucial problem for the safe deployment of machine learning models identifying samples that fall outside of the training distribution, i.e in-distribution data (\ID).
Most \OOD works focus on the classification models trained with Cross Entropy~(CE) and attempt to fix its inherent issues.
In this work we leverage powerful representation learned with Supervised Contrastive (SupCon) training and propose a holistic approach to learn a classifier robust to \OOD data.
We extend SupCon loss  with two additional contrast terms.
The first term pushes auxiliary \OOD representations away from \ID representations without imposing any constraints on similarities among auxiliary data.
The second term pushes \OOD features far from the existing class prototypes, while pushing \ID representations closer to their corresponding class prototype.
When auxiliary \OOD data is not available, we propose feature mixing techniques to efficiently generate pseudo-\OOD features.
Our solution is simple and efficient and acts as a natural extension of the closed-set supervised contrastive representation learning.
We compare against different~\OOD detection methods on the common benchmarks and show state-of-the-art results. 
\end{abstract}

%%%%%%%%% BODY TEXT
%%%%%%%%% BODY TEXT

\vspace{-0.5cm}
\section{Introduction}\label{sec:intro}
\vspace{-0.2cm}
% - DL works in closed-set settings, not so much for OOD sample.
Modern deep learning architectures have demonstrated great generalization performance, surpassing human baselines on different tasks \cite{he2015delving,zhang2017alignedreid,buetti2019deep}. However, these models are often trained and evaluated in a closed-set setting, where both train and test sets are assumed to be drawn from the same distribution (\ie, in-distribution data).

When encountered with examples coming from any other distribution (\ie, out-of-distribution data),
these models tend to give predictions that are highly confident but not reliable~\cite{yang2021generalized}. In an autonomous driving scenario, OOD samples might include new object classes, road signs or traffic conditions that the model has not seen during training. Therefore, for such safety-critical applications, it is vital to detect the OOD samples, avoid making predictions on them and possibly ask for human intervention instead.
\begin{figure}[t]
    \centering
    \includegraphics[width=0.45\textwidth]{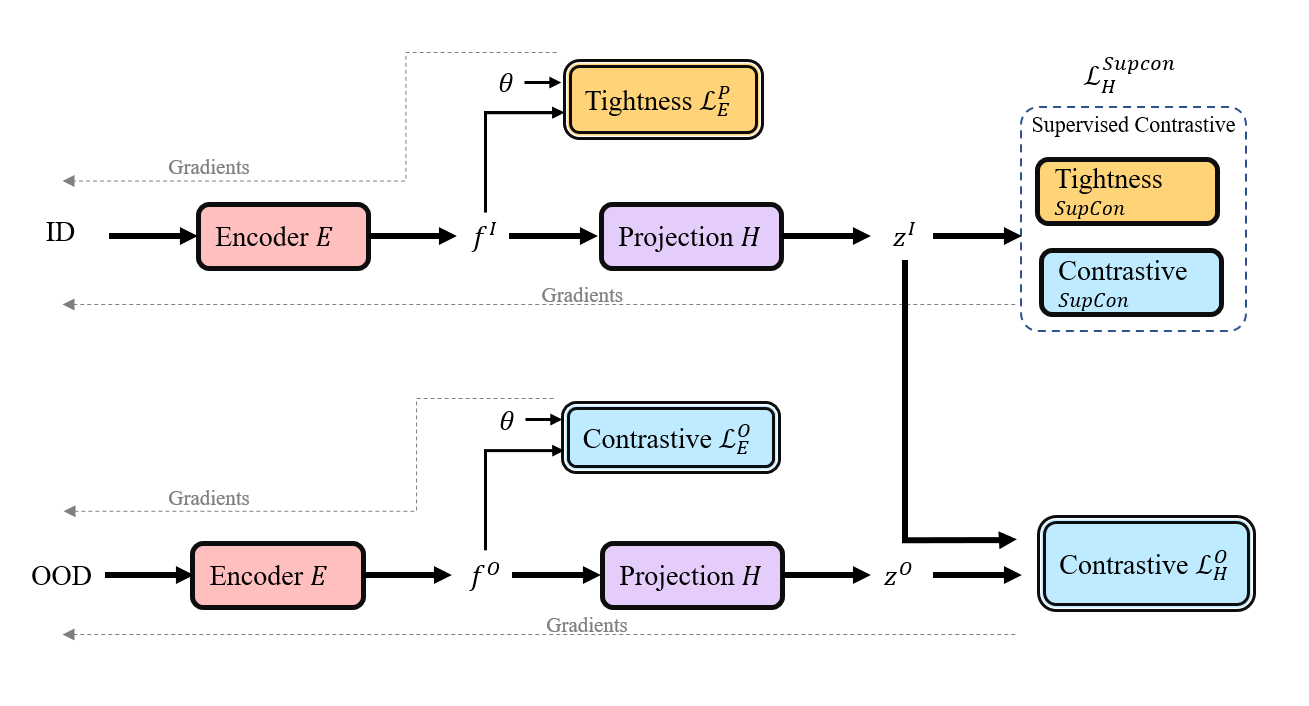}
    \vspace{-0.2cm}
    \caption{
    An illustration of our \OOD-aware Prototypical Supervised Contrastive Learning method. 
    We consider an encoder network extracting features from the input samples. The features are projected to a vector to which a supervised contrastive learning loss $\mathcal{L}^{\SupCon}$ is applied. Rather than using Cross-Entropy to learn a classifier on top of the features, we learn class prototypes $\th$ by applying a tightness term $\mathcal{L}_{E}^{P}$ to the in-distribution samples. This penalizes features that are far from others of the same class. In addition, we propose to use a contrastive term $\mathcal{L}_{H}^{O}$ to push apart projections coming from \ID and \OOD samples respectively. Likewise, we minimize the maximal similarity of \OOD features with the closest class prototypes using loss $\mathcal{L}_{E}^{O}$. The proposed new terms are marked with double outline.
    %\todo{Add \OOD samples pairs or change the whole concept of this figure}
    % An illustration of the additional positive (green) and negative (red) pairs proposed by our method.
    %The sample representations originating from the same class form positive pairs and are connected by the green arrows, while the negative pairs (of representations not from the same class) are connected by the red arrows.
    %Supervised Contrastive Loss~(SupCon) is used as the main feature representation learning method, where all possible pairs of samples are constructed.
    % Learned class prototypes are shown as squares with the same color as the corresponding class.
    %The class prototypes are shown as the squares of the color of the class they are learnt from.
    % Our loss constructs additional pairs while using prototypes as a proxy for ID classes.
    %We construct additional pairs using prototypes as the ID class proxies.
    %New positive pairs are formed between sample representations of a given class and the corresponding class prototype (pairwise similarities are maximized).
    %New negative pairs are formed between the representations of \OOD samples and the \ID classes prototypes (pairwise similarities are minimized). 
    }
    \label{fig:teaser}
    \vspace{-0.5cm}
\end{figure}
% - In classification, CE is the standard choice, but has drawbacks: it will likely produce high confidence given an OOD sample.

Cross-Entropy is a popular choice to train classification models under the closed-set assumption.
Popular datasets used in the closed-set setting \cite{deng2012mnist,deng2009imagenet,krizhevsky2009learning} have mutually exclusive classification labels that can be one-hot encoded.
A perfect fit for CE.
However, when a model is trained to always select an object class with a confidence close to $1$ for any input, it will likely produce highly confident predictions for~\OOD data as well ~\cite{wei2022mitigating}.
Besides, CE is shown to be sensitive to noise and susceptible to overfitting~\cite{berrada2018smooth}.

% - Arc: Other losses may have better properties.

% - An alternative to CE, SupCon may have other properties, more suitable to OOD detection.
Supervised contrastive training~\cite{khosla2020supervised} has been shown to improve the performance of a classification model by learning a rich representation of the samples.
The core idea is to leverage a large number of sample pairs and push sample representations of the same class to lie close together and far from the others in the embedding space.
This has been recently demonstrated to improve the~\OOD detection performance \cite{tack2020csi,winkens2020contrastive,wang2022partial} although it remains open how to explicitly employ the learned embedding for a better~\OOD detection.

Recent~\OOD detection methods, exposing the model to auxiliary~\OOD data during training, do not leverage the strength of representations learned with contrastive learning as they are tailored for the Softmax Cross-Entropy loss~\cite{hendrycks2018deep,liu2020energy}. In fact, jointly minimizing Softmax Cross-Entropy and Contrastive losses has been shown to lead to sub-optimal performance~\cite{khosla2020supervised}. 
% - Here is our hyphotesis: during training, use tightness and contrastive losses to ease OOD detection.

In this work, we propose an~\OOD-aware contrastive training objective.
We start from~\SupCon as a basis to learn the embedding.
Instead of relying on~\CE to learn the classifier weights, we learn prototypes, vectors lying in the same embedding as the feature extractor.
These prototypes are learnt by forming positive pairs of samples belonging to the same class and then maximizing their similarity.
We show that this prototype-based classifier provides less overconfident predictions on~\OOD data.
Next, we enrich~\SupCon with two loss terms that exploit any available auxiliary or synthesized~\OOD data.
The first loss term is applied at the projection head, similar to \SupCon, but targets minimizing the pairwise similarities of~\ID and~\OOD features.
Note that~\SupCon takes care of grouping~\ID features according to their classes while our first loss term pulls~\OOD features away from~\ID features.
The second loss term is applied at the feature extractor level and minimizes the likelihood of the~\OOD data, as per the prototype classifier, by pulling~\OOD features away from all the learned classes prototypes.
%In order to increase the \OOD robustness, we e abundant \OOD auxiliary data. During training, we encourage the model to project the \OOD features far from the \InDistribution class prototypes. Those \OOD features form negative pairs with the class samples prototypes on which similarities have to minimized. \todo{Negative pairs between prototypes and \OOD samples close to the class distribution in encoder/projection space?}
When auxiliary~\OOD data is unavailable, we propose~\ID features augmentation techniques to synthesize \OOD-like features leveraged to regularize the training.
Figure~\ref{fig:teaser} illustrates our proposed \OOD-aware contrastive training. 
% In this work, we train our network with a SupCon loss and define a prototype classifier \cite{yang2018robust} to replace CE with a tightness loss proposed in \cite{}. Our classifier learns a prototype for each category and pushes the samples to be close to their corresponding prototype in a contrastive scheme. Besides, we encourage the OOD samples to lie far from all class prototypes for a better OOD detection with an extra loss term. 
% Limitations of approaches which require past task data
% - We perform experiments to assess the performance of our model in several settings..., and present our contributions.

We evaluate our model in supervised and unsupervised settings where the~\OOD data is either available for training or it is synthesized using the available \ID~data.
Our model improves the~\OOD detection performance achieving state-of-the-art results, while maintaining or improving the classification accuracy on the \ID data.
In summary, the contributions of this work are:
\begin{itemize}
    % \item Starting from  supervised contrastive learning we propose to use a prototype-based prediction head instead of the CE trained linear head. 
    \item We propose an~\OOD-aware training scheme that, in combination with the representation learning loss, learns ID class prototypes and forces~\OOD features to lie further from the~\ID features and prototypes.
    \vspace{-0.2cm}
    \item When auxiliary \OOD~data is not available, we propose a simple and very efficient feature augmentation technique to generate \OOD-like features.
    \vspace{-0.2cm}
    \item Our experiments show the effectiveness of our training method compared to \CE-based models. We compare against existing \OOD works and show state-of-the-art results. We show an especially significant reduction in the false positive rate (FPR), an important metric in safety-critical applications.
    % \item We show that by using auxiliary \OOD data, either real or synthetically generated, we systematically improve the \OOD detection performance on other datasets, differently from recent \OOD detection methods that require millions of \OOD samples of diverse objects.
\end{itemize}

The rest of this paper is organized as follows.
We provide a background on existing methods in section~\ref{sec:background}. In section~\ref{sec:method} we detail our methodology and evaluate our proposed ideas in section~\ref{sec:experiments}.
We present conclusions in section~\ref{sec:conclusion}.
\vspace{-0.2cm}
\section{Related Work}\label{sec:background}
\vspace{-0.2cm}
% - Intro to OOD detection methods. Three kinds: 1) not changing model parameters, 2) ??? and 3) finetune using OOD data.
\OOD detection methods can be divided based on whether they operate on a fixed pretrained model, adapt the model parameters for the \OOD detection task, or leverage auxiliary \OOD data to fine-tune the model.

\textbf{Post-hoc methods} operate on the output of a pretrained model with different scoring functions for \OOD detection.
% There are many studies on OOD detection using the activations of a pretrained classifier without a need for extra training.
 Maximum Softmax Probability~\cite{hendrycks2016baseline} is among the most commonly used scoring functions.
 However, the softmax function is known to contribute to the highly confident predictions in DNNs and usually exhibits weak \OOD detection performance.
Liang~\cite{liang2017enhancing} proposed to enhance the separation of the softmax scores between \ID and \OOD inputs by temperature scaling and applying small perturbations to the input.
Similarly, with the goal of providing a more robust scoring function, a variety of different techniques were proposed, \eg, Mahalanobis distance to class centroids~\cite{lee2018simple}, predictions energy~\cite{liu2020energy} or maximum-logit~\cite{hendrycks2019scaling}.
These measures produce a wider range of confidence values compared to softmax and are easier to threshold for \OOD~detection.
 
More recently, building on the observation that \ID and \OOD inputs display highly distinctive signature patterns of network's internal activations, \cite{sun2021react} showed that clipping the activations of the penultimate layer of a pretrained model makes the output distributions for \ID and \OOD data better separated.
Targeting a similar phenomena, \cite{wei2022mitigating} proposed to normalize the logits to unit vectors before applying the CE loss during training instead, and showed strong \OOD detection performance. 
In this work, we use maximum-logit~\cite{hendrycks2019scaling} as our scoring function and overcome the logits-norm issue outlined in~\cite{wei2022mitigating} by operating only on normalized features and normalized prototypes during training.

%\textbf{Generative methods}
% - Generative models: OOD detection by high error in reconstruction
\textbf{Training-based methods} are broadly split into generative or self-supervised methods. 
Building on the assumption of an underlying distribution shift between \ID and \OOD data, there is a large body of literature on leveraging generative models \cite{kingma2013auto,van2016conditional,goodfellow2020generative} to capture the distribution of \ID examples and discard \OOD data \cite{bishop1994novelty,hendrycks2018deep,choi2018generative,nalisnick2019detecting,serra2019input,ren2019likelihood}.
Coupled with a generative model, reconstruction-based methods train an autoencoder on the existing \ID data and classify a sample as \OOD if the reconstruction error is high \cite{schlegl2017unsupervised,zong2018deep,perera2019ocgan}. Generative models remain however difficult to train and optimize in large scale and~\cite{nalisnick2018deep} challenged some of the assumptions on their feasibility for \OOD detection.
%\textbf{Self-supervised methods}
% - SS: Augmentations and contrastive learning.

Lately, self-supervised methods improve the \OOD detection performance by training the model to predict geometric transformations applied on the \ID data~\cite{golan2018deep,bergman2020classification,hendrycks2019using}.
More recently, \cite{winkens2020contrastive,tack2020csi,sehwag2021ssd, cho2021masked} revealed that different variants of contrastive training~\cite{chen2020simple} on the \ID~data improves the \OOD detection performance. 

In this work, we propose an \OOD-aware Supervised Contrastive (OPSupCon) training approach combining the supervised contrastive loss with additional tightness/contrastive losses to increase the \OOD robustness.
% In this work we train our network with a Supervised Contrastive (SupCon) loss \cite{khosla2020supervised} and our prototype classifier \cite{yang2018robust} with the tightness (TT) loss proposed in \cite{} that pushes the ID examples close to the class prototypes and the synthesized/real OOD far from them. 
%\textbf{Outlier Exposure}

\textbf{\OOD-leveraging methods}. As an alternative to post-hoc approaches or training-based methods relying solely on \InDistribution data, more powerful \OOD detection can be obtained by explicitly leveraging auxiliary \OOD data. 
Such works use supervision from \OOD samples collected from another mutually exclusive large dataset \cite{hendrycks2018deep}. In~\cite{hendrycks2018deep} CE loss is applied on auxiliary \OOD data with a uniform target distribution. \cite{liu2020energy} fine-tunes the model to explicitly create an energy gap by assigning low energy values to \InDistribution and high energy values to \OOD training data. 

% - What if no OOD aux data is available?
When auxiliary \OOD data is not available, works have instead synthesized  outlier examples using \InDistribution data \eg, by applying strong augmentations~\cite{tack2020csi} or by sampling outliers assuming that features follow normal distribution in the penultimate layer~\cite{du2022towards}.
Adversarial Reciprocal Point Learning (ARPL)~\cite{chen2021adversarial} constructs reciprocal points modeling the empty space between clusters of different classes samples.
In addition to generating confusing and diverse samples, a training scheme with adversarial margin constraint on the reciprocal points is proposed. However, this method is complex and requires intense hyperparameter tuning.
%or using tricks such as feature mixup.
In this work we propose a generic training scheme that includes \OOD exemplars in the contrastive training scheme. Besides, we consider the case where no representative \OOD data is available and alternatively propose feature manipulation techniques for generating pseudo OOD data. 
\vspace{-0.2cm}
\section{Methodology}\label{sec:method}
\vspace{-0.2cm}

%copied from ESUPCON paper
% Consider the following: 1) a random data augmentation module that for each sample $\x$ generates two differently augmented samples, 2) a neural network encoder $f$ that maps an augmented input sample $\x$ to its  feature representation: $f(\x)=\z, \z\in \mathbb{R}^d$. 

We consider a neural network that encodes each sample $\x$ with an encoder $E$, acting as a feature extractor, $E(\x)=\f$.
The projection head (\eg, a Multi Layer Perceptron) $H(\f)=\z$, maps the encoder feature $\f$ into the corresponding projection head feature $\z$.
Eventually, the \SupCon ~\cite{khosla2020supervised} loss is used to train both networks.
Typically, an additional linear classifier is trained on top of the encoder features $\f$ using \CE.
However, in order to avoid its argued short-comings, we replace the \CE-based training with \textit{learning} randomly initialized class prototypes $\th$ using a \textit{tightness} term that penalizes features $\f$ falling far from others of the sample class.
In addition, \textit{contrastive} terms push \OOD samples far from \ID samples and their prototypes.

We consider the setting where pairs of datum are available, one being an \ID sample together with its label $D^I=(X^I_i,y_i)$, the other an auxiliary \OOD sample $D^O={X^O_i}$. In section \ref{subsec:aux} we extend this concept to the case where no auxiliary~\OOD data is available.

\subsection{Loss Terms} \label{subsec:losses}
\vspace{-0.2cm}
Here we provide details of each loss term. An illustration of our OOD-aware Contrastive Learning method can be found in Fig. \ref{fig:teaser}.
\vspace{-0.4cm}
\paragraph{Losses on \ID data:}
\textbf{\SupCon loss on the head features}.
The~\SupCon learning encourages samples of the same class to be pushed close together and pulled away from the samples of other classes.
For a given~\ID sample embedding $\z_i^I$, we consider the embeddings of all other samples in the batch $\z^I_p$, belonging to the same class, as a set of positives $P_i$.
The~\SupCon loss, given the embedding and its set of positives is:
\begin{equation}\label{eq:supcon}
\begin{split}
% &\ell_\SupCon(\zn_i, P_i)=\\
&\mathcal{L}_{H,i}^{\SupCon}(\z^I_i, P_i)=\\
&-\frac{1}{|P_i|}\sum_{\z^{I}_p\in P_i} \log\frac{\exp(\similarity(\z^I_i,\z^I_p)/\tau)}{\sum_{j\ne i}\exp(\similarity(\z^I_i,\z^I_j)/\tau)} =\\
&\frac{1}{|P_i|}\sum_{\z^I_p\in P_i} \left(\underbrace{-{(\z^{I\top}_i\z^I_p)}{/\tau} }_\text{Tightness}+\underbrace{ \log \sum_{j\ne i}\exp\left(({\z^{I\top}_i\z^I_j)}{/\tau}\right)}_\text{Contrast}\right),
\end{split}
\end{equation}
where the index $j$ iterates over all (original and augmented) samples.
The~\SupCon loss is expressed as the average of the loss defined on each positive pair, where (in this supervised setting) the positive pairs are formed of augmented views and other samples of the same class.
Note, that the~\SupCon loss can be expressed as a combination of a tightness and a contrast term where positive pairs similarities are maximized via the tightness term and negative pairs similarities are minimized with the contrast term.

The total~\SupCon loss is the mean of the losses for the $N^{I}$ \ID samples considered.

\begin{equation}
% \ell_\SupCon=\frac{1}{N_n}\sum_i^{N_n}\ell_\SupCon(\zn_i, P_i),
\mathcal{L}^{\SupCon}_H=\frac{1}{N^I}\sum_{i=1}^{N^I}\mathcal{L}_{H,i}^{\SupCon}. 
\end{equation}
%\underbrace{(x + 2)^3}_\text{text 1}

% \textbf{A tightness loss on the encoder features} serves to learn class prototype $\th_y$ by maximizing their similarities with in-class samples.
\textbf{A tightness loss on the encoder features} serves to learn class prototypes $\th_c$ (in the encoder feature space) by maximizing their similarities with the corresponding class features:
\begin{equation}
    % \ell_\text{tight}=\frac{1}{N_n}\sum_i^{N_n}\ell_\text{tt}(\fn_i,\theta_{y_i})=\underbrace{\frac{1}{N_n} \sum_i^{N_n} -\fn_i^\top\theta_{y_i}}_\text{Tightness}.
    \mathcal{L}^P_E=\frac{1}{N^I}\sum_{i=1}^{N^I}\mathcal{L}^{\text{tt}}(\f^{I}_i,\th_{y_i})=\underbrace{\frac{1}{N^I} \sum_{i=1}^{N^I} -\f^{I\top}_i\th_{y_i}}_\text{Tightness}.
    \label{eq:tt}
\end{equation}

% We assume that both the samples representations $\f_i$ and the classifier weights $\th_i$ are normalized vectors and that the classifier is linear with no bias term.
% Specifically:
% \begin{equation}
% \f_{i}^*=\frac{\f_i}{\| \f_i \| }, \,
% \theta_{i}^*=\frac{\theta_i}{\| \theta_i \| }.
% %\theta_{p,i}^*=H(\theta_i)
% \end{equation}
We assume that all sample features $\f_i$ and class prototypes $\th_k$ are normalized to have unit length ($\|\f^I_i\|=\|\th_k\|=1$) and that the classifier is linear with no bias term.

Note that the number of samples in~\eqref{eq:tt} might differ from $N^I$ (\eg, due to augmentation), in which case $N^I$ should be replaced by the corresponding number of samples.
With that assumption, we use a nearest prototype classifier \ie, assigning a test sample to the class of the nearest prototype in the feature space.
% In this formulation, \InDistribution data during training are forced to be close to samples of the same classes.
% At the same time, \ID data are forced to be far from other classes samples, while the prototypes of each class are learned such that they are closest to their class samples.
With this formulation, features of the same class are forced to become closer and each class prototype is learned as the closest to its class features. Besides, features of different classes are forced to become further apart.
A similar loss term was introduced in~\cite{ESUPCON2022} to train only the linear classifier (the prototypes) for supervised classification. However, as discussed below, we use the tightness term in our work to also train the encoder which enhances the robustness when \OOD data are present during training.
\vspace{-0.4cm}
\paragraph{Losses on \OOD data.}
Auxiliary \OOD data are additional samples that do not belong to the concerned task's distribution.
No other information such as specific class labels or samples similarities are either provided  or can be assumed. Here we try to answer the question on how to increase the robustness of Supervised Contrastive training against \OOD data without assuming any specific additional information about these auxiliary data. We propose two additional loss terms to be combined with the aforementioned losses on \ID data, one at the level
of the projection head and the other at the encoder level.\\
\textbf{Contrastive term on the projection head features}.
When the \SupCon loss is applied on \ID samples, it is composed of the tightness term operating on positive pairs and the contrast term operating on other pairs.
For \OOD samples, we do not want to impose any superficial similarity to any other sample, the target is simply to learn how to project those samples as far from \ID samples as possible.
We thus propose to only deploy a contrast or a pull term to pairs of \OOD/\ID samples:
\begin{equation}\label{eq:contrast_head}
    \mathcal{L}^O_H = \frac{1}{N^O}\sum_{i=1}^{N^O}\underbrace{\log \sum^{N^I}_{j=1}\exp\left(({\z^{O\top}_i\z^I_j)}{/\tau}\right)}_\text{Contrast}.
\end{equation}\\
\textbf{Contrastive term on the encoder features}.
% The prototypes are learned as representatives of \ID classes in the encoder feature space.
% The similarity of each \ID feature with its corresponding prototype is maximized.
% For an \OOD sample, we know that it does not belong to any of the known classes.
% We propose to instead minimize its similarity to  all of the existing class prototypes:
The \OOD samples can be considered as coming from a new (w.r.t.\xspace the $K$ known \ID classes) category. Similarly to the projection head case~\eqref{eq:contrast_head}, the contrastive term for \OOD data can be defined on the encoder feature level.
In this case, instead of using \textit{all} \ID features $\f_i^I$, only their class prototypes $\th_k$ are deployed:
\begin{equation}\label{eq:contrast_encoder}
%  \ell_\text{cont} = \frac{1}{N_o}\sum_i^{N_o}\underbrace{ \log \sum_{k}\exp\left(({\zo_i^\top\theta_p,k)}{/\tau}\right)}_\text{Contrast}
% \mathcal{L}^{O}_E = \frac{1}{N^{O}}\sum_i^{N^{O}}\underbrace{ \log \sum_{k}\exp\left(({\z^{O\top}_i\theta_k)}{/\tau}\right)}_\text{Contrast}
%  \label{eq:contrast}
\mathcal{L}^O_E=\frac{1}{N^O}\sum_{i=1}^{N^O}\underbrace{\frac{1}{K}\log \sum_{k=1}^{K}\exp\left(({\f^{O\top}_i\th_k)}{/\tau}\right)}_\text{Contrast}.
\end{equation}
% We propose to minimize the log sum exponential~(LSE) of each \OOD sample similarity to the prototypes.
% Note, that this loss is a smooth approximation of the sample-prototype maximum similarity function, which smoothly minimizes the similarity of the \OOD samples to the closest class prototype.
We propose to minimize the Log Sum Exponential~(LSE) of each \OOD feature similarity to all prototypes,
which would lead to the desired minimization of the \textit{maximal} similarity of the \OOD feature to the closest class prototype.
\\
\textbf{The objective function}.
Our training objective on both \ID and \OOD samples is composed of losses operating on both the projection head and the encoder level. 
At the projection head, we operate on pairwise similarities. At the encoder level, however, the prototype-based proxy-similarities are optimized.
\ID samples and their class prototypes are encouraged to be close together, whereas the similarities between \ID prototypes and \OOD samples are minimized.
We hypothesize that such treatment of the \OOD samples would generalize better to other \OOD data, as opposed to imposing a specific clustering of \OOD data.
\begin{equation}%\label{eq:toal_loss}
% \ell=\ell_\SupCon +\ell_\text{tight}+\ell_\text{cont} 
\mathcal{L}=\mathcal{L}_H^{\SupCon} +\gamma\mathcal{L}_H^O + \alpha\left(\mathcal{L}_E^P +
 \mathcal{L}_E^O\right),
 \label{eq:final_loss}
\end{equation}
where the parameters $\alpha$ and $\gamma$ control the contribution of the additional.
The minimization of our final loss  optimizes jointly the representations and the class prototypes while attempting at increasing the \OOD robustness by contrasting auxiliary \OOD features from both \ID samples and their class prototypes. 

\begin{table*}[t]
\footnotesize
\begin{center}
\resizebox{\textwidth}{!}{ 
\begin{tabular}{|l|l l l|l l l|l l l|l l l|l l l|l l l|l l l|}
\hline
Dataset/\Longunderstack{Method \\ Metrics} & \multicolumn{3}{c|}{\Longunderstack{CE \\ \\ FPR$\downarrow$ AUROC$\uparrow$  AUPR$\uparrow$ \\}} &  \multicolumn{3}{c|}{\Longunderstack{PSupCon \\ \\ FPR$\downarrow$ AUROC$\uparrow$  \AUPR$\uparrow$\\}} & \multicolumn{3}{c|}{\Longunderstack{CE + Energy \\ \\ FPR$\downarrow$ AUROC$\uparrow$  AUPR$\uparrow$ \\}} & \multicolumn{3}{c|}{\Longunderstack{PSupCon + Energy \\ \\ FPR$\downarrow$ AUROC$\uparrow$  \AUPR$\uparrow$\\}} & \multicolumn{3}{c|}{\Longunderstack{OPSupCon-R \\ \\ FPR$\downarrow$ AUROC$\uparrow$  \AUPR$\uparrow$\\}} & \multicolumn{3}{c|}{\Longunderstack{OPSupCon-P \\ \\ FPR$\downarrow$ AUROC$\uparrow$  \AUPR$\uparrow$\\}}
\\
\hline
\hline
\rowcolor{\fprcolor}
DTD & 25.01 & 95.02 & 98.81  
 & 14.09 & 97.44 & 99.44
&  7.83 & 98.43 & 99.67
 &  \textbf{2.71} & \textbf{99.43} & \textbf{99.87}
 &  4.95 & 99.04 & 99.80
 &  16.57 & 96.69 & 99.22
 \\

\rowcolor{\auroccolor}
SVHN & 3.08 & 99.19 & 99.84 
 & 3.16 & 99.39 & 99.87 
& 1.55 & 99.47 & 99.90
 & 1.92 & 99.57 & 99.91
 & \textbf{0.85} & \textbf{99.75} & \textbf{99.95}
 &  5.41 & 98.46 & 99.70

\\

\rowcolor{\fprcolor}
Places365 & 28.56 & 94.07 & 98.52 
& 26.96 & 94.88 & 98.79 
& 20.61 & 95.70 & 98.94 
& 36.85 & 92.11 & 98.07
& 21.17 & 95.63 & 98.91
 & \textbf{14.48} & \textbf{96.76} & \textbf{99.21}
\\

\rowcolor{\auroccolor}
LSUN-C & 12.10 & 97.68 & 99.54 
& 3.74 & 99.22 & 99.84 
& 5.28 & 98.75 & 99.75
& 64.98 & 81.89 & 95.42
 &  \textbf{1.33} & \textbf{99.60} & \textbf{99.92}
 &  2.39 & 99.34 & 99.87

\\

\rowcolor{\fprcolor}
LSUN-R & 8.98 & 98.15 & 99.63 
& 6.43 & 98.65 & 99.73 
& 8.69 & 98.37 & 99.67 
& \textbf{4.40} & \textbf{99.11} & \textbf{99.81}
 &  9.52 & 98.16 & 99.64
 &  6.62 & 98.57 & 99.72

\\

\rowcolor{\auroccolor}
iSUN & 11.54 & 97.86 & 99.58 
& 6.29 & 98.71 & 99.74 
& 7.24 & 98.59 & 99.72 
& \textbf{2.48} & \textbf{99.44} & \textbf{99.88}
 & 7.71 & 98.40 & 99.69
 & 7.24 & 98.52 & 99.70
\\

\rowcolor{\fprcolor}
iNaturalist & 37.24 & 94.10 & 98.77 
 & 10.70 & 98.18 & 99.63 
& 18.49 & 96.40 & 99.21
 & \textbf{7.53} & \textbf{98.61} & \textbf{99.70} 
 &  9.87 & 98.11 & 99.63
 &   12.48 & 97.70 & 99.53
\\

\rowcolor{\auroccolor}
CIFAR-100 & 40.73 & 91.85 & 98.03 
 & 41.03 & 92.30 & 98.19 
& 37.04 & 93.00 & 98.34
 & 51.07 & 89.59 & 97.57
 &  36.42 & \textbf{93.25} & \textbf{98.51}
 &  \textbf{36.04} & 93.15 & 98.41
\\

\rowcolor{\fprcolor}
Mnist & 30.88 & 95.76 & 99.17 
 & \textbf{1.62} & \textbf{99.50} & \textbf{99.90} 
& 32.55 & 94.93 & 98.97
 & 45.78 & 92.85 & 98.53
 &  2.79 & 99.42 & 99.89
 &  8.10 & 98.55 & 99.72
\\

\rowcolor{\auroccolor}
TIN & 32.05 & 93.22 & 98.30 
 & 30.95 & 93.86 & 98.50 
& 27.00 & 94.36 & 98.58
 &  31.80 & 93.45 & 98.47
 &  25.83 & 94.39 & 98.61
 &  \textbf{25.55} & \textbf{94.61} & \textbf{98.64}
\\ 

\rowcolor{\fprcolor}
Average & 23.02 & 95.69 & 99.02 
 & 14.49 & 97.20 & 99.36 
& 16.63 & 96.80 & 99.27
 & 24.95 & 94.60 & 98.72
 &  \textbf{12.01} & \textbf{97.56} & \textbf{99.44} 
 &  13.52 & 97.24 & 99.38
 \\ 
\hline
\end{tabular}}
\vspace{-0.2cm}
\caption { \textbf{OOD detection performance on Cifar-10:} a) comparison of CE and PSupCon (1, 2 columns) and, b) comparison of OOD training with our method compared to energy finetuning. Our method outperforms performance  energy finetuning even with pseudo OOD.}
\label{tab:baselines_cifar10}
\end{center}
\vspace{-0.6cm}
\end{table*}

\subsection{Pseudo-\OOD features generation} \label{subsec:aux}
\vspace{-0.2cm}
Our method leverages auxiliary \OOD data to improve the \OOD robustness of the learned model.
Here, we propose a simple alternative  that generates pseudo \OOD features when auxiliary \OOD data cannot be provided.
% We consider here the setting where \OOD data are unavailable. We hypothesize that, in this case, the \OOD detection performance will also benefit from constructing the class prototypes far from parts of the embedding space likely to be occupied by samples drawn from different distributions. 
We suggest to transform \ID features to produce pseudo-\OOD features that mimic   realistic and challenging OOD cases. 
In many real-life applications, such as autonomous driving, there is a high chance of encountering \OOD inputs that lie in between class categories in the embedding space.
For instance, a model which has not seen any examples belonging to the class ``Motorcyclist'', may assign an internal representation to such examples close to both the pedestrian and vehicle features.  
Based on the observation that \OOD data are commonly projected in between \ID clusters and in  areas where different \ID  classes overlap~\cite{pmlr-v199-aljundi22a}, our idea is to generate features spanning this space between \ID samples of different classes.   %and leverage additionally the projection head for the purpose of \OOD regularization.

Inspired by the Manifold Mixup~\cite{pmlr-v97-verma19a} technique, where Mixup~\cite{zhang2018mixup} is applied at feature level (any hidden state) instead of at the input images, we suggest to perform a Mixup of the \ID features extracted from the encoder being trained. 
%The effect of Manifold Mixup is  smoother constructed decision boundaries the labels of mixed features are
Differently from existing Mixup techniques, we consider the generated features as \OOD and apply our proposed loss using this \OOD features at the projection head level. %where our \OOD loss term applied at projection head learned to project such features far from \ID features leading to a better separated class clusters. Our \OOD loss term applied at the encoder level also regularize the learned prototypes to be  far those pseudo features. 
% We suggest to apply augmentation techniques at the feature extraction level and apply our contrast loss on the projection head to encourage the network to project features far from classes prototypes. 
% \SupCon loss is usually applied on top of the projection head after the main encoder~\eqref{eq:supcon}. In this work, we leverage additionally the projection head for the purpose of \OOD regularization. We suggest to apply augmentation techniques at the feature extraction level and apply our contrast loss on the projection head to encourage the network to project features far from classes prototypes. 
Given an \ID feature we generate a pseudo-\OOD feature:
\begin{equation}\label{eq:mixup}
 %\fo = \lambda*\fn_i + (1-\lambda)*\fn_{j:y_j\ne y_i}.
 % \f_i^{O} = \lambda\cdot\f^{I}_i + (1-\lambda)\cdot\f^{I}_{j} : \, j=\underset{j}{\mathrm{argmax}} ({\f^{I\top}_i}\f^{I}_j ) \, \, s.t. \,\, y_j\ne y_i,
 \f_i^{O} = \lambda\f^{I}_i + (1-\lambda)\f^{I}_{j},\, j=\underset{j,y_j\ne y_i}\argmax({\f^{I\top}_i}\f^{I}_j),
\end{equation}
where the new pseudo-\OOD sample $\f_i^{O}$ is a linear combination of the concerned \ID feature $\f_i^I$ and the most similar \ID feature $\f_i^I$ of a different class. The selection of the closest feature of a different class is to ensure that the generated \OOD feature indeed lies between two close ID samples, of different classes, and to avoid generating redundant and easy \OOD features.
 The $\lambda$ is drawn at each iteration from a normal distribution centered at $0.5$ with $0.3$ standard deviation.
 Here \ID features come from raw and augmented \ID samples. The proposed pseudo-\OOD features generation technique is extremely efficient and adds minimal  computational cost. Further, as the pseudo features are generated at the encoder level, we can remove the term $\mathcal{L}_E^O$ from our full loss function (\ref{eq:final_loss}) and rely on $\mathcal{L}_H^O$ (\ref{eq:contrast_encoder}) applied at the projection head to train the model.

 %Note that here, differently from existing manifold mixup technique, we treat the resulting mixed features as if they were drawn from an unknown \OOD distribution.
% To further boost the coverage of generated features we stochastically retain pseudo-\OOD features from previous representations updates and combine them with the pseudo-\OOD features generated from the current representation.

% Once pseudo-\OOD features are constructed, they are forwarded through the projection head on which loss~\eqref{eq:contrast_head} is applied. 
% Once pseudo-\OOD features are constructed, they are used to train the projection head using the $\mathcal{L}_H^O$ loss~\eqref{eq:contrast_head}.
%The generated pseudo-\OOD features are added to the training set, which is used to train the projection head using the $\mathcal{L}_H^O$ loss~\eqref{eq:contrast_head}.

\vspace{-0.2cm}
\section{Experiments}\label{sec:experiments}
\vspace{-0.2cm}
In this section, we evaluate the effectiveness of our method and its components and compare it to state-of-the-art OOD detection methods on various datasets.
% In this section we evaluate  discuss the experimental setting and detail our approach for selecting or synthesizing \OOD training datasets. We further provide empirical evaluation on different types of classifiers, the main components of our contrastive training and compare our proposed method against various state-of-the-art methods. 

\vspace{-0.3cm}
\subsection{Experimental Settings}
\vspace{-0.2cm}
% In order to have a fair comparison with the previous literature, we adapt our settings to the unified scheme proposed in \cite{yang2022openood}.
%In particular, all methods reported in this section 
We employ a ResNet18~\cite{he2016deep} backbone, following  \cite{yang2022openood},  and use CIFAR-10 and CIFAR-100 \cite{krizhevsky2009learning} as the in-distribution datasets. We train our models with a batch size of 512  using SGD as the optimizer and a cosine annealing scheduler \cite{loshchilov2016sgdr}.  We use the same data augmentation as in \SupCon~\cite{khosla2020supervised}, namely AutoAugment. Training is performed for 500 epochs.
In the supplementary material we provide further results, using ResNet50 as the feature extractor. 
%\textbf{OOD dataset for training:}
%Throughout the experiments, we
% The Tiny Images dataset \cite{le2015tiny} is used whenever there is a requirement for real \OOD examples during training to which we remove any class intersection with either CIFAR-10 or CIFAR-100.  In the supplement we provide additional results by training our method with different \OOD data.

We extensively evaluate and report our results on Describable Textures Dataset (DTD) \cite{cimpoi2014describing}, SVHN \cite{netzer2011reading}, Places365 \cite{zhou2017places}, LSUN-Crop, LSUN-Resize \cite{yu2015lsun}, iSUN \cite{xu2015turkergaze}, iNaturalist \cite{van2018inaturalist}, Mnist \cite{deng2012mnist} and Tiny Imagenet (TIN) \cite{le2015tiny} and CIFAR datasets. %Additionally, we evaluate our method on CIFAR-10 and CIFAR-100 datasets for the experiments that do not use the examples from these datasets during training time. 
We use the following metrics to evaluate our experiments. %We provide a brief overview of each  dataset as well as the metrics used to evaluate the methods in the sequel.

\noindent\textbf{FPR@95 ($\downarrow$)},  measures the false positive rate when true positive rate is set to $95\%$, and referred to as \FPR. 

\noindent\textbf{\AUROC ($\uparrow$)},  the area under the Receiver Operating Characteristic (ROC) curve; denoting TPR/FPR relationship.

\noindent\textbf{\AUPR ($\uparrow$)}, the area under the Precision-Recall (PR) curve.
We consider \ID samples as positives.

\begin{table*}[t]
\footnotesize
\begin{center}
\resizebox{\textwidth}{!}{ 
\begin{tabular}{|l|l l l|l l l|l l l|l l l|l l l|l l l|l l l|}
\hline
Dataset/\Longunderstack{Method \\ Metrics} & \multicolumn{3}{c|}{\Longunderstack{CE \\ \\ FPR$\downarrow$ AUROC$\uparrow$  AUPR$\uparrow$ \\}} & \multicolumn{3}{c|}{\Longunderstack{PSupCon \\ \\ FPR$\downarrow$ AUROC$\uparrow$  \AUPR$\uparrow$\\}} & \multicolumn{3}{c|}{\Longunderstack{CE + Energy \\ \\ FPR$\downarrow$ AUROC$\uparrow$  AUPR$\uparrow$ \\}}  & \multicolumn{3}{c|}{\Longunderstack{PSupCon + Energy \\ \\ FPR$\downarrow$ AUROC$\uparrow$  \AUPR$\uparrow$\\}}  & \multicolumn{3}{c|}{\Longunderstack{OPSupCon-R \\ \\ FPR$\downarrow$ AUROC$\uparrow$  \AUPR$\uparrow$\\}} & \multicolumn{3}{c|}{\Longunderstack{OPSupCon-P \\ \\ FPR$\downarrow$ AUROC$\uparrow$  \AUPR$\uparrow$\\}}
\\
\hline
\hline
\rowcolor{\fprcolor}
DTD & 73.53 & 82.29 & 95.81
 & 66.14 & 80.14 & 94.53
 &   57.44 & 88.57 & 97.41
 &   \textbf{27.65} & \textbf{93.75} & \textbf{98.44}
&  51.22  &  88.44  &  97.28
 &  54.23 & 84.77 & 95.89

 \\

\rowcolor{\auroccolor}
SVHN & 34.38 & 93.89 & 98.71
 & 47.74 & 91.22 & 98.12
 & \textbf{17.21} & \textbf{96.93} & \textbf{99.34} 
 & 29.13 & 94.94 & 98.93
&  44.26 & 92.39 & 98.39
 &  49.49 & 90.89 & 98.04
\\

\rowcolor{\fprcolor}
Places365 & 80.83 & 77.96 & 94.52 
& 76.89 & 78.24 & 94.59
& 78.94 & 78.81 & 94.79 
 & 75.86 & 78.82 & 94.56
&  74.52 & 79.30 & 94.79
 &  \textbf{74.45} & \textbf{79.71} & \textbf{94.95}
\\

\rowcolor{\auroccolor}
LSUN-C & 54.84 & 90.19 & 97.93 
& 27.64 & 95.03 & 98.92
& 53.70 & 91.81 & 98.33 
 & 52.26 & 91.62 & 98.25
&  20.38 & 96.48 & 99.27
 &  \textbf{18.10} & \textbf{96.71} & \textbf{99.30}
\\

\rowcolor{\fprcolor}
LSUN-R & 62.42 & 88.00 & 97.42 
& 47.64 & 90.54 & 97.92
& 46.04 & 91.41 & 98.10 
 & \textbf{13.46} & \textbf{96.98} & \textbf{99.28}
&  38.54 & 93.01 & 98.49
 &  37.85 & 92.78 & 98.43
\\

\rowcolor{\auroccolor}
iSUN & 64.57 & 87.38 & 97.29 
& 55.10 & 88.29 & 97.38
& 50.00 & 90.33 & 97.85 
 & \textbf{14.38} & \textbf{96.92} & \textbf{99.29}
 & 46.45 & 91.33 & 98.13
 & 46.38 & 90.82 & 97.97
\\

\rowcolor{\fprcolor}
iNaturalist & 85.23 & 79.19 & 95.33 
& \textbf{43.67} & 89.11 & 97.24
& 79.15 & 83.25 & 96.32 
 & 70.28 & 85.19 & 96.71
& 47.71 & 89.87 & 97.63
 &  45.38 & \textbf{89.97} & \textbf{97.64}
\\

\rowcolor{\auroccolor}
CIFAR-10 & \textbf{76.84} & \textbf{79.16} & \textbf{94.88} 
& 83.45 & 73.12 & 92.66 
& 80.03 & 78.40 & 94.70 
& 89.84 & 71.60 & 92.75
&  84.74 & 71.01 & 91.50
 &  84.08 & 73.11 & 92.73
\\

\rowcolor{\fprcolor}
Mnist & 88.81 & 75.01 & 94.34 
& 33.93 & 94.24 & 98.79
& 95.33 & 67.34 & 92.36 
 & 96.27 & 77.41 & 95.21
&  33.89 & \textbf{94.38} & \textbf{98.83}
 &  \textbf{33.78} & 94.37 & \textbf{98.83}
\\

\rowcolor{\auroccolor}
TIN & 75.20 & 80.56 & 95.08 
& 72.15 & 81.09 & 95.22
& 71.70 & 82.91 & 95.68 
 &  \textbf{62.50} & \textbf{85.02} & \textbf{96.17}
&  68.00 & 82.67 & 95.52
 &  69.23 & 82.12 & 95.44
\\ 

\rowcolor{\fprcolor}
Average & 69.67 & 83.36 & 96.13 
& 55.43 & 86.10 & 96.53 
& 62.95 & 84.98 & 96.49
& 53.16 & 87.22 & 96.96 
&  \textbf{50.97} & \textbf{87.89} & \textbf{96.98}
&  51.29 & 87.53 & 96.92
 \\ 
\hline
\end{tabular}}
\vspace{-0.2cm}
\caption { \textbf{OOD detection performance on Cifar-100:} a) comparison of CE and PSupCon (1, 2 columns) and, b) comparison of OOD training with our method compared to energy finetuning. Our method outperforms performance  energy finetuning even with pseudo OOD.}
\label{tab:baselines_cifar100}
\vspace*{-0.6cm}
\end{center}
\end{table*}
\subsection{OOD Training Dataset}
\vspace{-0.2cm}
The selection of the dataset to be used as the \OOD data highly depends on whether it can comprehensively represent all other possible \OOD data or not, which in turn would depend on the distribution of each \ID dataset.
Many works such as ~\cite{hendrycks2018deep, liu2020energy} opt for a large diverse dataset, \ie, TinyImages \cite{torralba200880} which acts as an extensive set of all other possible objects. Such extensive \OOD dataset would likely present overlaps with many \ID datasets, and would require careful curation each time. Additionally, this dataset is not publicly available anymore due to ethical related issues. 

We use instead a more scalable approach: the much smaller DTD dataset~\cite{cimpoi2014describing}, a collection of various textures that can be synthetically generated using state-of-the-art generation techniques. This way, we show that our proposed contrastive training scheme can generalize to other \OOD datasets without accessing a huge collection of various types of \OOD objects. We further ablate the effect of the choice of auxiliary \OOD dataset in the supplementary material.

Moreover, as we have seen, an alternative to using a real \OOD dataset is to synthesize the auxiliary examples from the accessible \ID dataset \cite{tack2020csi, chen2021adversarial, huang2021mos, du2022vos, wei2022mitigating}. Similarly, we provide a simple alternative strategy to mimic \OOD like features as described in Section~\ref{subsec:aux}. We show that, using our method, the generalization performance is close to that of using a real \OOD dataset.
\vspace{-0.2cm}
\subsection{Compared methods}\label{sec:compared_methods}
\vspace{-0.2cm}
We compare the proposed method with representative state-of the-art methods.

\myparagraph{Post hoc methods.}
\noindent\texttt{MSP}~\cite{hendrycks2016baseline} uses the maximum Softmax probability as a scoring function, a standard baseline in \OOD detection literature. 
\noindent \texttt{ODIN}~\cite{liang2017enhancing} performs perturbation in the input image and uses the MSP score on the perturbed image.
\noindent \texttt{ReAct}~\cite{sun2021react} performs a rectification on the logits for computing the \OOD detection score (energy score). 

\myparagraph{Training-based methods.}
\noindent\texttt{CSI}~\cite{tack2020csi}, close to our work, leverages supervised contrastive learning. However, as a proxy for \OOD data, the method leverages strongly augmented samples. SSD \cite{sehwag2021ssd} combines contrastive learning on ID data with k-means clustering for OOD detection.
\noindent\texttt{ARPL}~\cite{Chen2021} proposes the concept of reciprocal points as representatives for \OOD data, and train the neural network such that the features of \ID classes lie within a margin distance from those points. 
\noindent\texttt{VOS}~\cite{du2022vos} synthesizes outliers in the penultimate layer, by assuming that \ID features follow a normal distribution within each class.
\noindent \texttt{LNorm}~\cite{wei2022mitigating} trains the network such that the logits norm is constrained to be a constant. 

\myparagraph{\OOD-leveraging methods.}
\noindent\texttt{OE}~\cite{hendrycks2018deep} makes use of \CE for \ID data.
For \OOD, they set an uniform distribution as the target. 
\noindent\texttt{UDG}~\cite{yang2021semantically} leverages unsupervised data for both \OOD training and enhancing \ID performance.
Two heads are proposed.
The first one minimizes \CE loss on \ID labeled data and maximizes the entropy on \OOD data.
The second performs deep clustering.

\noindent\texttt{Energy}~\cite{liu2020energy} maximizes an energy gap between the ID and OOD samples. First, the mean energy values on the ID (\textit{$m_{in}$}) and OOD (\textit{$m_{out}$}) datasets are calculated.  Next, the model is fine-tuned to produce energy values lower than \textit{$m_{in}$} for the ID samples and higher than \textit{$m_{out}$} for OOD samples. It achieves the best OOD detection performance compared to previous OOD training methods. However, an  extra step to calculate the thresholds \textit{$m_{in}$} and \textit{$m_{out}$} is required.

\myparagraph{Our method.}
We refer to our method as \ours: OOD-aware Prototypical Supervised Contrastive learning.
Throughout the experiments we consider different variants of our method.
\textbf{\SupTT} refers to the combination of supervised contrastive training loss~\eqref{eq:supcon} with the prototypes learning loss~\eqref{eq:tt}, \OOD regularization is not applied here. 
\textbf{\oursreal} refers to models trained with our complete loss~\eqref{eq:final_loss} based on \textbf{r}eal auxiliary \OOD data.
\textbf{\oursfake} refers to models trained with $\mathcal{L}_H^O$ in loss~\eqref{eq:final_loss}. We generate \textbf{p}seudo \OOD-like features from \ID examples using~\eqref{eq:mixup}, as described in Section~\ref{subsec:aux}.

We use Maximum Logit~\cite{hendrycks2019scaling} as our scoring function.
Logit here refers to the  dot product between a sample representation and a given prototype. We ablate the choice of different scoring functions in the supplementary materials.

%We further ablate different variations of our method by combining them with the finetuning scheme defined in \cite{liu2020energy}. We denote those experiments with an extra \textit{(Energy)} term added to the method's name.

\subsection{\SupTT \OOD detection performance}
\vspace{-0.2cm}
We first compare the \OOD detection performance of the prototype classifier, trained with \SupCon~\cite{khosla2020supervised} and the tightness term~(\SupTT), to a classifier trained with a Cross-Entropy~(\CE) loss. The purpose is to observe the inherent \OOD detection capability of each model without explicit OOD fine-tuning.

Detailed results can be found in tables~\ref{tab:baselines_cifar10} and~\ref{tab:baselines_cifar100} (columns 1 and 2). 
%To achieve the best performance we have trained all the models in this table for 500 epochs.
We see that \SupTT consistently outperforms \CE on most datasets and metrics with a a significant reduction on the \FPR. This is a especially relevant metric for the purpose of rejecting \OOD samples, as it is calculated where the rejection rate is fixed to $5\%$. Achieving a lower \FPR is crucial for real life applications, where it is not possible to know the threshold in advance.

Therefore, the prototype classifier based on SupCon is shown to be more robust for \OOD detection. % We further show in the supplementary material that it achieves a favorable classification accuracy compared to \CE as well.

\vspace{-0.2cm}
\subsection{OOD-Aware Supervised Contrastive Learning}
\vspace{-0.2cm}
The formulation of our proposed loss function (\ref{eq:final_loss}) permits its different components to be applied at different stages, as required by different use cases. We first train our model with SupCon~\cite{khosla2020supervised} and fine-tune it with our full objective function~\eqref{eq:final_loss} for additional $50$~epochs.
This allows learning a good initial representation of the task at hand and improving these representation for a stronger separation of ID and OOD data. It also permits fine-tuning of any pre-trained model when \OOD data becomes available.

In our loss function ~\eqref{eq:final_loss}, we use $\gamma=1$ for training with real~\OOD and $\gamma=0.5$ for synthesized~\OOD.
The weight of the encoder losses $(\mathcal{L}_E^P +
 \mathcal{L}_E^O)$ is set to $\alpha=0.1$.

\vspace{-0.2cm}
\subsubsection{Comparison with SOTA \OOD Training method}
\vspace{-0.2cm}
First we extensively compare our proposed OOD training scheme with energy fine-tuning\cite{liu2020energy}. This method shows state-of-the-art performance when fine-tuning a pretrained model with real auxiliary OOD data. Energy fine-tuning was originally introduced for models trained with Cross-Entropy loss. Here we compare our proposed method with energy fine-tuning on top of both \CE and \SupTT  models.

Tables~\ref{tab:baselines_cifar10} and~\ref{tab:baselines_cifar100} summarise our results for this purpose for Cifar-10 and Cifar-100 datasets respectively. We observe that \oursreal and \oursfake (columns 5 and 6) outperform the models fine-tuned with Energy~\cite{liu2020energy} (columns 3 and 4) on most datasets, achieving a better average for all metrics. Moreover, energy fine-tuning on top of a \SupTT model improves the results on some datasets while significantly worsening the results on some others, proving {\textit{ unreliable}} for OOD detection.

Note that Energy fine-tuning~\cite{liu2020energy} requires an extra step to determine  thresholds \textit{$m_{in}$} and \textit{$m_{out}$} for each model, before fine-tuning. Our method achieves a better performance without any extra model-dependant hyper-parameters. %To the best our knowledge, our method is the only OOD training method designed to leverage the power of SupCon~\cite{khosla2020supervised}

\begin{table}[h!]
\scriptsize
\centering
\setlength{\extrarowheight}{0pt}
\addtolength{\extrarowheight}{\aboverulesep}
\addtolength{\extrarowheight}{\belowrulesep}
\setlength{\aboverulesep}{0pt}
\setlength{\belowrulesep}{0pt}
\begin{tabular}{|p{1.1cm}|p{0.6cm}|p{\metricsize}p{\metricsize}p{\metricsize}p{\metricsize}p{\metricsize}p{\metricsize}p{\metricsize}|} 
\toprule
Method                                           & Metric      & \begin{sideways}DTD\end{sideways}        & \multicolumn{1}{|l|}{\begin{sideways}SVHN\end{sideways}} & \multicolumn{1}{l|}{\begin{sideways}Places365\end{sideways}} & \multicolumn{1}{l|}{\begin{sideways}CIFAR-100\end{sideways}} & \multicolumn{1}{l|}{\begin{sideways}MNIST\end{sideways}} & \multicolumn{1}{l|}{\begin{sideways}TIN\end{sideways}} & \begin{sideways}Average\end{sideways}     \\ 
\hline
\hline
\multirow{3}{*}{\Longunderstack{CE + Energy}} & {\cellcolor{\fprcolor}}{\tiny FPR$\downarrow$}   & {\cellcolor{\fprcolor}} 39.45 & {\cellcolor{\fprcolor}} 20.41               & {\cellcolor{\fprcolor}} 34.12               & {\cellcolor{\fprcolor}} 60.42                    & {\cellcolor{\fprcolor}} 51.02              & {\cellcolor{\fprcolor}} 49.75           & {\cellcolor{\fprcolor}} 42.52 \\
                                                 & {\cellcolor{\auroccolor}}{\tiny AUROC$\uparrow$} & {\cellcolor{\auroccolor}}  93.61  & {\cellcolor{\auroccolor}} 96.78                 & {\cellcolor{\auroccolor}} 92.97                        & {\cellcolor{\auroccolor}}  88.41                   & {\cellcolor{\auroccolor}} 93.36                 & {\cellcolor{\auroccolor}} 90.47                & {\cellcolor{\auroccolor}} 92.60 \\
                                                 & {\cellcolor{\auprcolor}}{\tiny AUPR$\uparrow$}  & {\cellcolor{\auprcolor}} 98.68 & {\cellcolor{\auprcolor}} 99.37                 & {\cellcolor{\auprcolor}} 98.37                     & {\cellcolor{\auprcolor}} 97.40                       & {\cellcolor{\auprcolor}}  98.73                   & {\cellcolor{\auprcolor}} 97.79              & {\cellcolor{\auprcolor}} 98.05 \\ 
\hline
\multirow{3}{*}{\Longunderstack{\ours\\R}} & {\cellcolor{\fprcolor}}{\tiny FPR$\downarrow$}   & {\cellcolor{\fprcolor}} \textbf{8.27} & {\cellcolor{\fprcolor}} \textbf{3.27}               & {\cellcolor{\fprcolor}} \textbf{21.98}                     & {\cellcolor{\fprcolor}} \textbf{43.70}                     & {\cellcolor{\fprcolor}} 6.46                & {\cellcolor{\fprcolor}} \textbf{33.12}               & {\cellcolor{\fprcolor}} \textbf{19.46}  \\
                                                 & {\cellcolor{\auroccolor}}{\tiny AUROC$\uparrow$} & {\cellcolor{\auroccolor}}  \textbf{98.48} & {\cellcolor{\auroccolor}} \textbf{99.26}                & {\cellcolor{\auroccolor}} \textbf{95.37}                     & {\cellcolor{\auroccolor}}  \textbf{91.20}                      & {\cellcolor{\auroccolor}} 98.58                & {\cellcolor{\auroccolor}} 93.40               & {\cellcolor{\auroccolor}} \textbf{96.04}  \\
                                                 & {\cellcolor{\auprcolor}}{\tiny AUPR$\uparrow$}  & {\cellcolor{\auprcolor}} \textbf{99.68} & {\cellcolor{\auprcolor}} \textbf{99.85}                & {\cellcolor{\auprcolor}} \textbf{98.83}                     & {\cellcolor{\auprcolor}} \textbf{97.87}                    & {\cellcolor{\auprcolor}} 99.72                & {\cellcolor{\auprcolor}} \textbf{98.36}             & {\cellcolor{\auprcolor}} \textbf{99.05}  \\ 
\hline
\multirow{3}{*}{\Longunderstack{\ours\\P}} & {\cellcolor{\fprcolor}}{\tiny FPR$\downarrow$}   & {\cellcolor{\fprcolor}} 18.65 & {\cellcolor{\fprcolor}} 4.88                & {\cellcolor{\fprcolor}} 25.02                     & {\cellcolor{\fprcolor}} 46.43                     & {\cellcolor{\fprcolor}} \textbf{4.48}                  & {\cellcolor{\fprcolor}} 34.23              & {\cellcolor{\fprcolor}} 22.28   \\
                                                 & {\cellcolor{\auroccolor}}{\tiny AUROC$\uparrow$} & {\cellcolor{\auroccolor}} 96.11 & {\cellcolor{\auroccolor}}  99.00                & {\cellcolor{\auroccolor}}  95.00                    & {\cellcolor{\auroccolor}} 90.48                    & {\cellcolor{\auroccolor}}   \textbf{98.97}                   & {\cellcolor{\auroccolor}} 93.16              & {\cellcolor{\auroccolor}} 95.45  \\
                                                 & {\cellcolor{\auprcolor}}{\tiny AUPR$\uparrow$}  & {\cellcolor{\auprcolor}} 99.07 & {\cellcolor{\auprcolor}} 99.80                & {\cellcolor{\auprcolor}} 98.79                   & {\cellcolor{\auprcolor}} 97.78                    & {\cellcolor{\auprcolor}} \textbf{99.80}                 & {\cellcolor{\auprcolor}} 98.30              & {\cellcolor{\auprcolor}}98.92  \\ 
\hline
\multirow{3}{*}{MSP~\cite{hendrycks2016baseline}}                             & {\cellcolor{\fprcolor}}{\tiny FPR$\downarrow$}   & {\cellcolor{\fprcolor}}59.89 & {\cellcolor{\fprcolor}}51.87                & {\cellcolor{\fprcolor}}57.64                     & {\cellcolor{\fprcolor}}62.01                     & {\cellcolor{\fprcolor}}58.59                 & {\cellcolor{\fprcolor}}60.69               & {\cellcolor{\fprcolor}}58.44  \\
                                                 & {\cellcolor{\auroccolor}}{\tiny AUROC$\uparrow$} & {\cellcolor{\auroccolor}} 88.72 & {\cellcolor{\auroccolor}}90.88                & {\cellcolor{\auroccolor}}89.03                     & {\cellcolor{\auroccolor}}87.11                     & {\cellcolor{\auroccolor}}89.91                 & {\cellcolor{\auroccolor}}86.62                & {\cellcolor{\auroccolor}} 88.71 \\
                                                 & {\cellcolor{\auprcolor}}{\tiny AUPR$\uparrow$}  & {\cellcolor{\auprcolor}}91.28 & {\cellcolor{\auprcolor}}78.19                & {\cellcolor{\auprcolor}}70.24                     & {\cellcolor{\auprcolor}}85.92                     & {\cellcolor{\auprcolor}}66.95                 & {\cellcolor{\auprcolor}}83.07               & {\cellcolor{\auprcolor}}79.27  \\ 
\hline
\multirow{3}{*}{ODIN~\cite{liang2017enhancing}}                            & {\cellcolor{\fprcolor}}{\tiny FPR$\downarrow$}   & {\cellcolor{\fprcolor}}51.10 & {\cellcolor{\fprcolor}}67.92                & {\cellcolor{\fprcolor}}50.51                     & {\cellcolor{\fprcolor}}59.09                     & {\cellcolor{\fprcolor}}36.23                 & {\cellcolor{\fprcolor}}59.06               & {\cellcolor{\fprcolor}}53.98  \\
                                                 & {\cellcolor{\auroccolor}}{\tiny AUROC$\uparrow$} & {\cellcolor{\auroccolor}}80.70 & {\cellcolor{\auroccolor}}73.32                & {\cellcolor{\auroccolor}}82.55                     & {\cellcolor{\auroccolor}}77.68                     & {\cellcolor{\auroccolor}}90.91                 & {\cellcolor{\auroccolor}} 77.33               & {\cellcolor{\auroccolor}} 80.41  \\
                                                 & {\cellcolor{\auprcolor}}{\tiny AUPR$\uparrow$}  & {\cellcolor{\auprcolor}}82.25 & {\cellcolor{\auprcolor}}42.13                & {\cellcolor{\auprcolor}}50.27                     & {\cellcolor{\auprcolor}}73.24                     & {\cellcolor{\auprcolor}}64.74                 & {\cellcolor{\auprcolor}}70.07               & {\cellcolor{\auprcolor}}63.78  \\ 
\hline
\multirow{3}{*}{ReAct~\cite{sun2021react}}                           & {\cellcolor{\fprcolor}}{\tiny FPR$\downarrow$}   & {\cellcolor{\fprcolor}}49.98 & {\cellcolor{\fprcolor}}49.23                & {\cellcolor{\fprcolor}}44.21                     & {\cellcolor{\fprcolor}}53.72                     & {\cellcolor{\fprcolor}}50.94                 & {\cellcolor{\fprcolor}}47.00               & {\cellcolor{\fprcolor}}47.68  \\
                                                 & {\cellcolor{\auroccolor}}{\tiny AUROC$\uparrow$} & {\cellcolor{\auroccolor}}88.18 & {\cellcolor{\auroccolor}}89.50                & {\cellcolor{\auroccolor}}90.09                     & {\cellcolor{\auroccolor}}86.35                     & {\cellcolor{\auroccolor}}88.34                 & {\cellcolor{\auroccolor}}88.90               & {\cellcolor{\auroccolor}}88.56  \\
                                                 & {\cellcolor{\auprcolor}}{\tiny AUPR$\uparrow$}  & {\cellcolor{\auprcolor}}89.91 & {\cellcolor{\auprcolor}}75.36                & {\cellcolor{\auprcolor}}69.28                     & {\cellcolor{\auprcolor}} 83.15                     & {\cellcolor{\auprcolor}}50.88                 & {\cellcolor{\auprcolor}}86.53               & {\cellcolor{\auprcolor}}75.85  \\ 
\hline
\multirow{3}{*}{CSI~\cite{tack2020csi}}                             & {\cellcolor{\fprcolor}}{\tiny FPR$\downarrow$}   & {\cellcolor{\fprcolor}}53.63 & {\cellcolor{\fprcolor}}33.26                & {\cellcolor{\fprcolor}}58.01                     & {\cellcolor{\fprcolor}}61.92                     & {\cellcolor{\fprcolor}}32.07                 & {\cellcolor{\fprcolor}}55.27               & {\cellcolor{\fprcolor}}49.02  \\
                                                 & {\cellcolor{\auroccolor}}{\tiny AUROC$\uparrow$} & {\cellcolor{\auroccolor}}91.04 & {\cellcolor{\auroccolor}}95.22                & {\cellcolor{\auroccolor}}88.57                     & {\cellcolor{\auroccolor}}88.08                     & {\cellcolor{\auroccolor}}95.09                 & {\cellcolor{\auroccolor}}90.18               & {\cellcolor{\auroccolor}}91.36  \\
                                                 & {\cellcolor{\auprcolor}}{\tiny AUPR$\uparrow$}  & {\cellcolor{\auprcolor}}95.00 & {\cellcolor{\auprcolor}}92.42                & {\cellcolor{\auprcolor}}77.57                     & {\cellcolor{\auprcolor}}89.87                     & {\cellcolor{\auprcolor}}86.30                 & {\cellcolor{\auprcolor}}92.12               & {\cellcolor{\auprcolor}}88.88  \\ 
\hline
\multirow{3}{*}{ARPL~\cite{chen2021adversarial}}                            & {\cellcolor{\fprcolor}}{\tiny FPR$\downarrow$}   & {\cellcolor{\fprcolor}}69.86 & {\cellcolor{\fprcolor}}73.41                & {\cellcolor{\fprcolor}}66.20                     & {\cellcolor{\fprcolor}}69.81                     & {\cellcolor{\fprcolor}}68.99                 & {\cellcolor{\fprcolor}}68.46               & {\cellcolor{\fprcolor}}69.45  \\
                                                 & {\cellcolor{\auroccolor}}{\tiny AUROC$\uparrow$} & {\cellcolor{\auroccolor}}87.36 & {\cellcolor{\auroccolor}}87.77                & {\cellcolor{\auroccolor}}88.40                     & {\cellcolor{\auroccolor}}86.68                     & {\cellcolor{\auroccolor}}88.48                  & {\cellcolor{\auroccolor}}87.70               & {\cellcolor{\auroccolor}}87.73  \\
                                                 & {\cellcolor{\auprcolor}}{\tiny AUPR$\uparrow$}  & {\cellcolor{\auprcolor}}92.85 & {\cellcolor{\auprcolor}}82.92                & {\cellcolor{\auprcolor}}77.63                     & {\cellcolor{\auprcolor}}88.65                     & {\cellcolor{\auprcolor}}74.27                 & {\cellcolor{\auprcolor}}89.87               & {\cellcolor{\auprcolor}}84.36  \\ 
\hline
\multirow{3}{*}{VOS~\cite{du2022vos}}                             & {\cellcolor{\fprcolor}}{\tiny FPR$\downarrow$}   & {\cellcolor{\fprcolor}}37.38 & {\cellcolor{\fprcolor}}29.92                & {\cellcolor{\fprcolor}}45.37                     & {\cellcolor{\fprcolor}}52.94                     & {\cellcolor{\fprcolor}}42.22                 & {\cellcolor{\fprcolor}}45.85               & {\cellcolor{\fprcolor}}42.28  \\
                                                 & {\cellcolor{\auroccolor}}{\tiny AUROC$\uparrow$} & {\cellcolor{\auroccolor}}91.26 & {\cellcolor{\auroccolor}}93.82                & {\cellcolor{\auroccolor}}88.73                     & {\cellcolor{\auroccolor}}86.08                     & {\cellcolor{\auroccolor}}89.83                 & {\cellcolor{\auroccolor}}88.89               & {\cellcolor{\auroccolor}}89.76  \\
                                                 & {\cellcolor{\auprcolor}}{\tiny AUPR$\uparrow$}  & {\cellcolor{\auprcolor}}92.72 & {\cellcolor{\auprcolor}}83.73                & {\cellcolor{\auprcolor}}63.93                     & {\cellcolor{\auprcolor}}83.52                     & {\cellcolor{\auprcolor}}52.37                 & {\cellcolor{\auprcolor}}87.15               & {\cellcolor{\auprcolor}}77.23  \\ 
\hline
\multirow{3}{*}{LNorm~\cite{wei2022mitigating}}                           & {\cellcolor{\fprcolor}}{\tiny FPR$\downarrow$}   & {\cellcolor{\fprcolor}}30.94 & {\cellcolor{\fprcolor}} 5.30                & {\cellcolor{\fprcolor}}31.17                     & {\cellcolor{\fprcolor}}46.99                     & {\cellcolor{\fprcolor}}4.75                 & {\cellcolor{\fprcolor}}36.34               & {\cellcolor{\fprcolor}}25.91  \\
                                                 & {\cellcolor{\auroccolor}}{\tiny AUROC$\uparrow$} & {\cellcolor{\auroccolor}}94.30 & {\cellcolor{\auroccolor}}98.86                & {\cellcolor{\auroccolor}}94.76                     & {\cellcolor{\auroccolor}}91.13                     & {\cellcolor{\auroccolor}}98.82                 & {\cellcolor{\auroccolor}}\textbf{93.90}               & {\cellcolor{\auroccolor}}95.29  \\
                                                 & {\cellcolor{\auprcolor}}{\tiny AUPR$\uparrow$}  & {\cellcolor{\auprcolor}}96.32 & {\cellcolor{\auprcolor}}97.70                & {\cellcolor{\auprcolor}}88.11                     & {\cellcolor{\auprcolor}}91.89                     & {\cellcolor{\auprcolor}}96.24                 & {\cellcolor{\auprcolor}}94.84               & {\cellcolor{\auprcolor}}94.18  \\ 
\hline
\multirow{3}{*}{OE~\cite{hendrycks2018deep}}                              & {\cellcolor{\fprcolor}}{\tiny FPR$\downarrow$}   & {\cellcolor{\fprcolor}}79.49 & {\cellcolor{\fprcolor}}84.59                & {\cellcolor{\fprcolor}}    84.69                 & {\cellcolor{\fprcolor}}    82.14                  & {\cellcolor{\fprcolor}}  94.32               & {\cellcolor{\fprcolor}}  78.44            & {\cellcolor{\fprcolor}}83.90  \\
                                                 & {\cellcolor{\auroccolor}}{\tiny AUROC$\uparrow$} & {\cellcolor{\auroccolor}}78.90 & {\cellcolor{\auroccolor}}82.40                & {\cellcolor{\auroccolor}}72.06                     & {\cellcolor{\auroccolor}}75.35                     & {\cellcolor{\auroccolor}}67.31                 & {\cellcolor{\auroccolor}}77.37               & {\cellcolor{\auroccolor}}75.56  \\
                                                 & {\cellcolor{\auprcolor}}{\tiny AUPR$\uparrow$}  & {\cellcolor{\auprcolor}}85.78 & {\cellcolor{\auprcolor}}73.96                & {\cellcolor{\auprcolor}}41.59                     & {\cellcolor{\auprcolor}}75.35                     & {\cellcolor{\auprcolor}}35.09                 & {\cellcolor{\auprcolor}}77.80               & {\cellcolor{\auprcolor}}65.22  \\
\hline
\multirow{3}{*}{UDG~\cite{yang2021semantically}}                              & {\cellcolor{\fprcolor}}{\tiny FPR$\downarrow$}   & {\cellcolor{\fprcolor}}43.97 & {\cellcolor{\fprcolor}}61.91                & {\cellcolor{\fprcolor}}42.44                     & {\cellcolor{\fprcolor}}55.33                     & {\cellcolor{\fprcolor}}39.32                 & {\cellcolor{\fprcolor}}42.48               & {\cellcolor{\fprcolor}}47.57  \\
                                                 & {\cellcolor{\auroccolor}}{\tiny AUROC$\uparrow$} & {\cellcolor{\auroccolor}}93.56 & {\cellcolor{\auroccolor}}92.50                & {\cellcolor{\auroccolor}}93.58                     & {\cellcolor{\auroccolor}}90.38                     & {\cellcolor{\auroccolor}}93.81                 & {\cellcolor{\auroccolor}}93.33               & {\cellcolor{\auroccolor}}92.86  \\
                                                 & {\cellcolor{\auprcolor}}{\tiny AUPR$\uparrow$}  & {\cellcolor{\auprcolor}}96.55 & {\cellcolor{\auprcolor}}90.85                & {\cellcolor{\auprcolor}}87.89                     & {\cellcolor{\auprcolor}}91.67                     & {\cellcolor{\auprcolor}}82.67                 & {\cellcolor{\auprcolor}}94.66               & {\cellcolor{\auprcolor}}90.71  \\

\hline
\multirow{3}{*}{\Longunderstack{SSD \\SupCon~\cite{sehwag2021ssd}}}                              & {\cellcolor{\fprcolor}}{\tiny FPR$\downarrow$}   & {\cellcolor{\fprcolor}} 24.29 & {\cellcolor{\fprcolor}}\textbf{1.67}                & {\cellcolor{\fprcolor}}29.52                     & {\cellcolor{\fprcolor}}49.18                      & {\cellcolor{\fprcolor}}8.38                 & {\cellcolor{\fprcolor}}44.06               & {\cellcolor{\fprcolor}}26.18  \\
                                                 & {\cellcolor{\auroccolor}}{\tiny AUROC$\uparrow$} & {\cellcolor{\auroccolor}}95.97 & {\cellcolor{\auroccolor}}\textbf{99.65}                & {\cellcolor{\auroccolor}}94.09                      & {\cellcolor{\auroccolor}}89.34                     & {\cellcolor{\auroccolor}} 98.13                  & {\cellcolor{\auroccolor}}90.52               & {\cellcolor{\auroccolor}}94.61  \\
                                                 & {\cellcolor{\auprcolor}}{\tiny AUPR$\uparrow$}  & {\cellcolor{\auprcolor}}93.12 & {\cellcolor{\auprcolor}}\textbf{99.86}                & {\cellcolor{\auprcolor}}\textbf{99.76}                    & {\cellcolor{\auprcolor}}88.43                     & {\cellcolor{\auprcolor}}97.23                 & {\cellcolor{\auprcolor}}89.77               & {\cellcolor{\auprcolor}}94.69  \\

\bottomrule
\end{tabular}
\vspace*{-0.2cm}
\caption {\footnotesize\textbf{Comparison with the state-of-the-art on CIFAR-10 dataset.} \ours improves significantly over state of the art methods.}
%For each method we report three metrics: \FPR$\downarrow$, \AUROC$\uparrow$ and \AUPR$\uparrow$.}
\label{tab:sota_comparison_cifar10}
\vspace*{-0.5cm}
\end{table}

\begin{table}
\vspace*{-0.2cm}
\scriptsize
\centering
\setlength{\extrarowheight}{0pt}
\addtolength{\extrarowheight}{\aboverulesep}
\addtolength{\extrarowheight}{\belowrulesep}
\setlength{\aboverulesep}{0pt}
\setlength{\belowrulesep}{0pt}
\begin{tabular}{|p{1.1cm}|p{0.6cm}|p{\metricsize}p{\metricsize}p{\metricsize}p{\metricsize}p{\metricsize}p{\metricsize}p{\metricsize}|} 
\toprule
Method                                           & Metric      & \begin{sideways}DTD\end{sideways}        & \multicolumn{1}{|l|}{\begin{sideways}SVHN\end{sideways}} & \multicolumn{1}{l|}{\begin{sideways}Places365\end{sideways}} & \multicolumn{1}{l|}{\begin{sideways}CIFAR-10\end{sideways}} & \multicolumn{1}{l|}{\begin{sideways}MNIST\end{sideways}} & \multicolumn{1}{l|}{\begin{sideways}TIN\end{sideways}} & \begin{sideways}Average\end{sideways}     \\ 
\hline
\hline
\multirow{3}{*}{\Longunderstack{CE + Energy}} & {\cellcolor{\fprcolor}}{\tiny FPR$\downarrow$}   & {\cellcolor{\fprcolor}} 77.45 & {\cellcolor{\fprcolor}} \textbf{24.79}               & {\cellcolor{\fprcolor}} 76.25                  & {\cellcolor{\fprcolor}} 87.67                     & {\cellcolor{\fprcolor}} 93.71                  & {\cellcolor{\fprcolor}} 74.45               & {\cellcolor{\fprcolor}} 72.38  \\
                                                 & {\cellcolor{\auroccolor}}{\tiny AUROC$\uparrow$} & {\cellcolor{\auroccolor}}   77.66  & {\cellcolor{\auroccolor}}\textbf{95.28}               & {\cellcolor{\auroccolor}} 75.77                    & {\cellcolor{\auroccolor}}  71.05                      & {\cellcolor{\auroccolor}} 65.95                & {\cellcolor{\auroccolor}} 79.71            & {\cellcolor{\auroccolor}} 77.57  \\
                                                 & {\cellcolor{\auprcolor}}{\tiny AUPR$\uparrow$}  & {\cellcolor{\auprcolor}} 94.24 & {\cellcolor{\auprcolor}} \textbf{98.96}                   & {\cellcolor{\auprcolor}} 93.60                       & {\cellcolor{\auprcolor}} 92.23                      & {\cellcolor{\auprcolor}}  92.06                & {\cellcolor{\auprcolor}} 94.89             & {\cellcolor{\auprcolor}} 94.33  \\ 
\hline
\multirow{3}{*}{\Longunderstack{\ours\\R}} & {\cellcolor{\fprcolor}}{\tiny FPR$\downarrow$}   & {\cellcolor{\fprcolor}} \textbf{43.54} & {\cellcolor{\fprcolor}} 79.73              & {\cellcolor{\fprcolor}} 77.59                     & {\cellcolor{\fprcolor}} 87.21                     & {\cellcolor{\fprcolor}} \textbf{9.75}                 & {\cellcolor{\fprcolor}} \textbf{73.63}              & {\cellcolor{\fprcolor}} \textbf{61.90}  \\
                                                 & {\cellcolor{\auroccolor}}{\tiny AUROC$\uparrow$} & {\cellcolor{\auroccolor}} \textbf{90.81} & {\cellcolor{\auroccolor}} 83.84                & {\cellcolor{\auroccolor}} 78.70                     & {\cellcolor{\auroccolor}}70.69                      & {\cellcolor{\auroccolor}} \textbf{98.52}                 & {\cellcolor{\auroccolor}} 80.98               & {\cellcolor{\auroccolor}} \textbf{83.92}  \\
                                                 & {\cellcolor{\auprcolor}}{\tiny AUPR$\uparrow$}  & {\cellcolor{\auprcolor}} \textbf{97.78} & {\cellcolor{\auprcolor}} 96.59                & {\cellcolor{\auprcolor}} 94.61                     & {\cellcolor{\auprcolor}} 91.83                    & {\cellcolor{\auprcolor}}  \textbf{99.70}               & {\cellcolor{\auprcolor}} \textbf{95.15}              & {\cellcolor{\auprcolor}} \textbf{95.94}  \\ 
\hline
\multirow{3}{*}{\Longunderstack{\ours\\P}} & {\cellcolor{\fprcolor}}{\tiny FPR$\downarrow$}   & {\cellcolor{\fprcolor}} 57.25  & {\cellcolor{\fprcolor}} 83.20               & {\cellcolor{\fprcolor}} 76.75                     & {\cellcolor{\fprcolor}} 84.70                     & {\cellcolor{\fprcolor}} 20.61                  & {\cellcolor{\fprcolor}} 74.12              & {\cellcolor{\fprcolor}} 66.10  \\
                                                 & {\cellcolor{\auroccolor}}{\tiny AUROC$\uparrow$} & {\cellcolor{\auroccolor}} 84.19  & {\cellcolor{\auroccolor}} 82.16                & {\cellcolor{\auroccolor}}  78.94                       & {\cellcolor{\auroccolor}} 73.49                     & {\cellcolor{\auroccolor}} 96.69                 & {\cellcolor{\auroccolor}} 80.60               & {\cellcolor{\auroccolor}} 82.67  \\
                                                 & {\cellcolor{\auprcolor}}{\tiny AUPR$\uparrow$}  & {\cellcolor{\auprcolor}} 95.71 & {\cellcolor{\auprcolor}} 96.20               & {\cellcolor{\auprcolor}} 94.68                     & {\cellcolor{\auprcolor}} \textbf{92.97}                     & {\cellcolor{\auprcolor}} 99.30                & {\cellcolor{\auprcolor}} 95.00               & {\cellcolor{\auprcolor}} 95.64  \\ 
\hline
\multirow{3}{*}{MSP~\cite{hendrycks2016baseline}}                             & {\cellcolor{\fprcolor}}{\tiny FPR$\downarrow$}   & {\cellcolor{\fprcolor}}83.83 & {\cellcolor{\fprcolor}}83.69                & {\cellcolor{\fprcolor}}81.24                     & {\cellcolor{\fprcolor}}\textbf{81.82}                     & {\cellcolor{\fprcolor}}87.78                 & {\cellcolor{\fprcolor}}76.22               & {\cellcolor{\fprcolor}}82.43  \\
                                                 & {\cellcolor{\auroccolor}}{\tiny AUROC$\uparrow$} & {\cellcolor{\auroccolor}}76.93 & {\cellcolor{\auroccolor}}76.04                & {\cellcolor{\auroccolor}}79.44                     & {\cellcolor{\auroccolor}}\textbf{78.31}                     & {\cellcolor{\auroccolor}}77.78                 & {\cellcolor{\auroccolor}}81.78               & {\cellcolor{\auroccolor}}78.38  \\
                                                 & {\cellcolor{\auprcolor}}{\tiny AUPR$\uparrow$}  & {\cellcolor{\auprcolor}}85.24 & {\cellcolor{\auprcolor}}60.76                & {\cellcolor{\auprcolor}}62.39                     & {\cellcolor{\auprcolor}}79.58                     & {\cellcolor{\auprcolor}}54.19                 & {\cellcolor{\auprcolor}}86.30               & {\cellcolor{\auprcolor}}71.41  \\ 
\hline
\multirow{3}{*}{ODIN~\cite{liang2017enhancing}}                            & {\cellcolor{\fprcolor}}{\tiny FPR$\downarrow$}   & {\cellcolor{\fprcolor}}83.83 & {\cellcolor{\fprcolor}}83.69                & {\cellcolor{\fprcolor}}81.27                     & {\cellcolor{\fprcolor}}83.16                     & {\cellcolor{\fprcolor}}75.34                 & {\cellcolor{\fprcolor}}77.77               & {\cellcolor{\fprcolor}}80.84  \\
                                                 & {\cellcolor{\auroccolor}}{\tiny AUROC$\uparrow$} & {\cellcolor{\auroccolor}}79.39 & {\cellcolor{\auroccolor}}71.08                & {\cellcolor{\auroccolor}}79.83                     & {\cellcolor{\auroccolor}}78.18                     & {\cellcolor{\auroccolor}}83.71                 & {\cellcolor{\auroccolor}}81.39               & {\cellcolor{\auroccolor}}78.93  \\
                                                 & {\cellcolor{\auprcolor}}{\tiny AUPR$\uparrow$}  & {\cellcolor{\auprcolor}}86.67 & {\cellcolor{\auprcolor}}52.36                & {\cellcolor{\auprcolor}}60.85                     & {\cellcolor{\auprcolor}}79.12                     & {\cellcolor{\auprcolor}}62.02                 & {\cellcolor{\auprcolor}}85.30               & {\cellcolor{\auprcolor}}71.05  \\ 
\hline
\multirow{3}{*}{ReAct~\cite{sun2021react}}                           & {\cellcolor{\fprcolor}}{\tiny FPR$\downarrow$}   & {\cellcolor{\fprcolor}}76.76 & {\cellcolor{\fprcolor}}77.41                & {\cellcolor{\fprcolor}}79.18                    & {\cellcolor{\fprcolor}}82.89                     & {\cellcolor{\fprcolor}}89.32                 & {\cellcolor{\fprcolor}}75.81               & {\cellcolor{\fprcolor}}80.22  \\
                                                 & {\cellcolor{\auroccolor}}{\tiny AUROC$\uparrow$} & {\cellcolor{\auroccolor}}{81.73} & {\cellcolor{\auroccolor}}83.73                & {\cellcolor{\auroccolor}}79.63                     & {\cellcolor{\auroccolor}}76.98                     & {\cellcolor{\auroccolor}}77.02                 & {\cellcolor{\auroccolor}}\textbf{81.96}               & {\cellcolor{\auroccolor}}80.17  \\
                                                 & {\cellcolor{\auprcolor}}{\tiny AUPR$\uparrow$}  & {\cellcolor{\auprcolor}}{89.01} & {\cellcolor{\auprcolor}}76.43                & {\cellcolor{\auprcolor}}59.44                     & {\cellcolor{\auprcolor}}77.78                     & {\cellcolor{\auprcolor}}52.01                 & {\cellcolor{\auprcolor}}85.89               & {\cellcolor{\auprcolor}}73.42  \\ 
\hline
\multirow{3}{*}{CSI~\cite{tack2020csi}}                             & {\cellcolor{\fprcolor}}{\tiny FPR$\downarrow$}   & {\cellcolor{\fprcolor}}89.27 & {\cellcolor{\fprcolor}}67.96                & {\cellcolor{\fprcolor}}87.91                     & {\cellcolor{\fprcolor}}88.23                     & {\cellcolor{\fprcolor}}92.38                 & {\cellcolor{\fprcolor}}85.30               & {\cellcolor{\fprcolor}}85.17  \\
                                                 & {\cellcolor{\auroccolor}}{\tiny AUROC$\uparrow$} & {\cellcolor{\auroccolor}}59.72 & {\cellcolor{\auroccolor}}78.57                & {\cellcolor{\auroccolor}}69.94                     & {\cellcolor{\auroccolor}}69.24                     & {\cellcolor{\auroccolor}}57.06                 & {\cellcolor{\auroccolor}}72.32               & {\cellcolor{\auroccolor}}67.80  \\
                                                 & {\cellcolor{\auprcolor}}{\tiny AUPR$\uparrow$}  & {\cellcolor{\auprcolor}}68.86 & {\cellcolor{\auprcolor}}60.24                & {\cellcolor{\auprcolor}}48.53                     & {\cellcolor{\auprcolor}}71.03                     & {\cellcolor{\auprcolor}}27.43                 & {\cellcolor{\auprcolor}}78.18               & {\cellcolor{\auprcolor}}59.04  \\ 
\hline
\multirow{3}{*}{ARPL~\cite{chen2021adversarial}}                            & {\cellcolor{\fprcolor}}{\tiny FPR$\downarrow$}   & {\cellcolor{\fprcolor}}88.76 & {\cellcolor{\fprcolor}}80.90                & {\cellcolor{\fprcolor}}85.25                     & {\cellcolor{\fprcolor}}85.80                     & {\cellcolor{\fprcolor}}84.91                 & {\cellcolor{\fprcolor}}83.34               & {\cellcolor{\fprcolor}}84.82  \\
                                                 & {\cellcolor{\auroccolor}}{\tiny AUROC$\uparrow$} & {\cellcolor{\auroccolor}}69.50 & {\cellcolor{\auroccolor}}78.97                & {\cellcolor{\auroccolor}}74.57                    & {\cellcolor{\auroccolor}}73.48                     & {\cellcolor{\auroccolor}}72.94                 & {\cellcolor{\auroccolor}}76.31               & {\cellcolor{\auroccolor}}74.29  \\
                                                 & {\cellcolor{\auprcolor}}{\tiny AUPR$\uparrow$}  & {\cellcolor{\auprcolor}}79.33 & {\cellcolor{\auprcolor}}68.58                & {\cellcolor{\auprcolor}}55.80                     & {\cellcolor{\auprcolor}}75.19                     & {\cellcolor{\auprcolor}}43.31                 & {\cellcolor{\auprcolor}}82.39               & {\cellcolor{\auprcolor}}67.43  \\ 
\hline
\multirow{3}{*}{VOS~\cite{du2022vos}}                             & {\cellcolor{\fprcolor}}{\tiny FPR$\downarrow$}   & {\cellcolor{\fprcolor}}94.54 & {\cellcolor{\fprcolor}}98.62                & {\cellcolor{\fprcolor}}97.81                     & {\cellcolor{\fprcolor}}96.64                     & {\cellcolor{\fprcolor}}92.31                 & {\cellcolor{\fprcolor}}96.40               & {\cellcolor{\fprcolor}}96.05  \\
                                                 & {\cellcolor{\auroccolor}}{\tiny AUROC$\uparrow$} & {\cellcolor{\auroccolor}}68.33 & {\cellcolor{\auroccolor}}68.99                & {\cellcolor{\auroccolor}}68.21                      & {\cellcolor{\auroccolor}}71.74                     & {\cellcolor{\auroccolor}}82.17                 & {\cellcolor{\auroccolor}}72.08               & {\cellcolor{\auroccolor}}{71.92}  \\
                                                 & {\cellcolor{\auprcolor}}{\tiny AUPR$\uparrow$}  & {\cellcolor{\auprcolor}}76.20 & {\cellcolor{\auprcolor}}56.36                & {\cellcolor{\auprcolor}}43.20                     & {\cellcolor{\auprcolor}}72.17                     & {\cellcolor{\auprcolor}}55.66                 & {\cellcolor{\auprcolor}}77.10               & {\cellcolor{\auprcolor}}63.44  \\ 
\hline
\multirow{3}{*}{LNorm~\cite{wei2022mitigating}}                           & {\cellcolor{\fprcolor}}{\tiny FPR$\downarrow$}   & {\cellcolor{\fprcolor}}87.06 & {\cellcolor{\fprcolor}}79.16                & {\cellcolor{\fprcolor}}80.20                     & {\cellcolor{\fprcolor}}83.77                     & {\cellcolor{\fprcolor}}{53.07}                 & {\cellcolor{\fprcolor}}77.19               & {\cellcolor{\fprcolor}}76.74  \\
                                                 & {\cellcolor{\auroccolor}}{\tiny AUROC$\uparrow$} & {\cellcolor{\auroccolor}}71.53 & {\cellcolor{\auroccolor}}83.03                & {\cellcolor{\auroccolor}}\textbf{79.84}                     & {\cellcolor{\auroccolor}}74.84                     & {\cellcolor{\auroccolor}}{90.82}                 & {\cellcolor{\auroccolor}}81.87               & {\cellcolor{\auroccolor}}80.32  \\
                                                 & {\cellcolor{\auprcolor}}{\tiny AUPR$\uparrow$}  & {\cellcolor{\auprcolor}}79.08 & {\cellcolor{\auprcolor}} 75.57                & {\cellcolor{\auprcolor}} 63.10                     & {\cellcolor{\auprcolor}}73.56                     & {\cellcolor{\auprcolor}}76.09                 & {\cellcolor{\auprcolor}}86.28               & {\cellcolor{\auprcolor}} 75.61  \\ 
\hline
\multirow{3}{*}{OE~\cite{hendrycks2018deep}}                              & {\cellcolor{\fprcolor}}{\tiny FPR$\downarrow$}   & {\cellcolor{\fprcolor}}88.46 & {\cellcolor{\fprcolor}}75.31                & {\cellcolor{\fprcolor}}92.23                     & {\cellcolor{\fprcolor}}90.92                     & {\cellcolor{\fprcolor}}80.84                 & {\cellcolor{\fprcolor}}90.85               & {\cellcolor{\fprcolor}}86.43  \\
                                                 & {\cellcolor{\auroccolor}}{\tiny AUROC$\uparrow$} & {\cellcolor{\auroccolor}}64.70 & {\cellcolor{\auroccolor}}77.43                & {\cellcolor{\auroccolor}}64.91                     & {\cellcolor{\auroccolor}}63.23                     & {\cellcolor{\auroccolor}}76.89                 & {\cellcolor{\auroccolor}}64.14               & {\cellcolor{\auroccolor}}68.55  \\
                                                 & {\cellcolor{\auprcolor}}{\tiny AUPR$\uparrow$}  & {\cellcolor{\auprcolor}}74.66 & {\cellcolor{\auprcolor}}62.15                & {\cellcolor{\auprcolor}}45.23                     & {\cellcolor{\auprcolor}}64.65                     & {\cellcolor{\auprcolor}}49.18                 & {\cellcolor{\auprcolor}}72.25               & {\cellcolor{\auprcolor}}61.35  \\
\hline
\multirow{3}{*}{UDG~\cite{yang2021semantically}}                              & {\cellcolor{\fprcolor}}{\tiny FPR$\downarrow$}   & {\cellcolor{\fprcolor}}83.46 & {\cellcolor{\fprcolor}}93.47                & {\cellcolor{\fprcolor}}\textbf{74.13}                     & {\cellcolor{\fprcolor}}87.22                     & {\cellcolor{\fprcolor}}93.28                 & {\cellcolor{\fprcolor}}78.21               & {\cellcolor{\fprcolor}} 84.96  \\
                                                 & {\cellcolor{\auroccolor}}{\tiny AUROC$\uparrow$} & {\cellcolor{\auroccolor}}72.15 & {\cellcolor{\auroccolor}}53.38               & {\cellcolor{\auroccolor}}78.61                     & {\cellcolor{\auroccolor}}72.88                     & {\cellcolor{\auroccolor}}66.63                 & {\cellcolor{\auroccolor}}78.79               & {\cellcolor{\auroccolor}}70.40  \\
                                                 & {\cellcolor{\auprcolor}}{\tiny AUPR$\uparrow$}  & {\cellcolor{\auprcolor}}81.32 & {\cellcolor{\auprcolor}}28.68                & {\cellcolor{\auprcolor}}57.34                     & {\cellcolor{\auprcolor}}75.00                     & {\cellcolor{\auprcolor}}32.54                 & {\cellcolor{\auprcolor}}83.32               & {\cellcolor{\auprcolor}}59.7  \\

\hline
\multirow{3}{*}{\Longunderstack{SSD \\SupCon~\cite{sehwag2021ssd}}}                           & {\cellcolor{\fprcolor}}{\tiny FPR$\downarrow$}   & {\cellcolor{\fprcolor}}48.10& {\cellcolor{\fprcolor}}28.46                & {\cellcolor{\fprcolor}}81.00                      & {\cellcolor{\fprcolor}}86.66                     & {\cellcolor{\fprcolor}}52.62                & {\cellcolor{\fprcolor}}76.22              & {\cellcolor{\fprcolor}} 62.17  \\
                                                 & {\cellcolor{\auroccolor}}{\tiny AUROC$\uparrow$} & {\cellcolor{\auroccolor}}90.59 & {\cellcolor{\auroccolor}}94.45               & {\cellcolor{\auroccolor}}76.46                     & {\cellcolor{\auroccolor}}66.45                     & {\cellcolor{\auroccolor}}89.26                 & {\cellcolor{\auroccolor}} 79.18               & {\cellcolor{\auroccolor}}82.73  \\
                                                 & {\cellcolor{\auprcolor}}{\tiny AUPR$\uparrow$}  & {\cellcolor{\auprcolor}}83.99 & {\cellcolor{\auprcolor}}98.70                & {\cellcolor{\auprcolor}}\textbf{98.83}                     & {\cellcolor{\auprcolor}}64.32                     & {\cellcolor{\auprcolor}}88.04                 & {\cellcolor{\auprcolor}}76.30               & {\cellcolor{\auprcolor}}85.03  \\

\bottomrule
\end{tabular}
\vspace*{-0.2cm}
\caption {\footnotesize\textbf{Comparison with the state-of-the-art on CIFAR-100 dataset.} Our  \ours improves significantly over state of the art methods.}
%For each method we report three metrics: \FPR$\downarrow$, \AUROC$\uparrow$ and \AUPR$\uparrow$}.
\label{tab:sota_comparison_cifar100}
\vspace*{-0.5cm}
\end{table}
\vspace{-0.4cm}
\subsubsection{Comparison with other methods}
\vspace{-0.2cm}
Tables \ref{tab:sota_comparison_cifar10} and \ref{tab:sota_comparison_cifar100} compare our method to different lines of literature described in Sec \ref{sec:compared_methods} on CIFAR-10 and CIFAR-100 datasets\footnote{The results for previous work (except for Energy and SSD) are taken from \cite{yang2022openood} and can be found \href{https://docs.google.com/spreadsheets/d/1gGHpdA3sSgfGpsrDUgt9lejvIbf5yOrDyJ511zsn1uY/edit?usp=sharing}{here}}.
We follow the evaluation protocol of \cite{yang2022openood} by training the model for 100 epochs only and using a ResNet-18 architecture. OOD fine-tuning is set to 10 epochs.

% We use Tinyimages (TIM) dataset for training with real \OOD data while taking out the indices that are similar to the examples in the CIFAR dataset.

% \begin{figure}
%      \centering
%      \begin{subfigure}[b]{0.15\textwidth}
%          \centering
%          \includegraphics[width=\textwidth]{images/placeholder.png}
%          \caption{TIM dog sample}
%          \label{fig:TIM-dog-sample}
%      \end{subfigure}
%      \begin{subfigure}[b]{0.15\textwidth}
%          \centering
%          \includegraphics[width=\textwidth]{images/placeholder.png}
%          \caption{TIM dog sample}
%          \label{fig:CIFAR-10-dog-sample}
%      \end{subfigure}
%      \begin{subfigure}[b]{0.15\textwidth}
%          \centering
%          \includegraphics[width=\textwidth]{images/placeholder.png}
%          \caption{TIM dog sample}
%          \label{fig:CIFAR-100-dog-sample}
%      \end{subfigure}
%         \caption{Example}
%         \label{fig:three graphs}
% \end{figure}

The values reported for previous work might slightly differ from those in the original papers due to the architecture change (ResNet-18) and decreased number of training epochs (100 epochs). Note that this is suboptimal for our approach as well since supervised contrastive training usually requires more epochs to converge. However, \cite{yang2022openood} enables the community to have a fair, complete and model independent comparison of different lines of work. 

Our \oursreal outperforms state-of-the-art methods from different families reducing the average \FPR rate by $22.81$ on CIFAR-10 and $10.48$ on CIFAR-100.
These results suggests the effectiveness of our training pipeline when leveraging real auxiliary \OOD data.

We further show with \oursfake that our training scheme can benefit from synthesized \OOD-like features and supersede the rival state-of-the-art methods, even those which leverage real-\OOD data, thanks to the powerful representation training mechanism. On CIFAR-100, SSD \cite{sehwag2021ssd} achieves an overall better performance for the FPR and AUROC metrics compared to \oursfake. While \oursfake does better on majority of the datasets, the difference on SVHN biases this comparison. More extensive experimentation on a larger number of datasets shows that our method achieves a better overal result to this work by a margin (supplementary material).

Finally, in the supplementary materials we show that further combining pseudo and real \OOD data can provide an extra boost to the detection performance.
%\oursfake achieves the best average AUROC on CIFAR-10 (96.57) and on CIFAR-100 (82.85)..
\vspace{-0.4cm}
\subsubsection{{Ablation}}
\vspace{-0.2cm}
In this section we evaluate the effectiveness of the different  components of our loss. We ablate this by fine-tuning the SupCon trained model with: \\
1) Prototype losses on the encoder level only ($\mathcal{L}_H^{\SupCon} +\alpha\mathcal{L}_E^P$).\\
2) Auxiliary loss on the head level combined with the SupCon loss and the prototype loss ($\mathcal{L}_H^{\SupCon} +\mathcal{L}_H^O +\alpha\mathcal{L}_E^P$).\\
3) Full loss function, with a contrast term applied at the encoder level as well ($\mathcal{L}_H^{\SupCon} +\gamma\mathcal{L}_H^O + \alpha\left(\mathcal{L}_E^P +  \mathcal{L}_E^O\right)$).\\

Table~\ref{tab:ablation} reports the results of each variant in the case of training with real auxiliary OOD data with Cifar-10 as the ID dataset. While fine-tuning with the prototypes loss alone does not bring substantial improvements, minimizing our auxiliary loss on the head does. Adding the extra contrastive term at encoder level further enhances the quality of the prototypes, increasing the robustness of the OOD detection.
\begin{table}[t!]
\vspace*{-0.2cm}
\footnotesize
\centering
\begin{center}
%\begin{tabular}{|l|l l l|l l l|l l l|l l l|l l l|}
\begin{tabular}{|l|l|l|l|}
\hline
Method & FPR$\downarrow$ &  AUROC$\uparrow$  & AUPR$\uparrow$ \\
\hline
\hline
\rowcolor{\fprcolor}
$\mathcal{L}_H^{\SupCon} +\alpha\mathcal{L}_E^P$& 14.46 & 97.14 & 99.35     \\
\rowcolor{\auroccolor}
%17.91  96.63  98.03
$\mathcal{L}_H^{\SupCon} +\mathcal{L}_H^O +\alpha\mathcal{L}_E^P$& 12.45 & 97.43 & 99.41  \\
%4.74  99.04  98.10

\rowcolor{\fprcolor}
$\mathcal{L}_H^{\SupCon} +\gamma\mathcal{L}_H^O + \alpha\left(\mathcal{L}_E^P +  \mathcal{L}_E^O\right)$&\textbf{12.01} &\textbf{97.56}& \textbf{99.44}    \\
%27.42 & 94.65 & 48.84
%12.01 & 97.56 & 99.44

%\mathcal{L}_H^{\SupCon} +\gamma\mathcal{L}_H^O + \alpha\left(\mathcal{L}_E^P +  \mathcal{L}_E^O\right)

%1.18  99.67  99.69
\hline
\end{tabular}
\vspace{-0.2cm}
\caption {\footnotesize\textbf{ Ablation study:} Investigating the effect of the  components of our loss function on CIFAR-10 dataset. Average mean results over the different OOD datasets are reported. }
\label{tab:ablation}
\vspace*{-0.4cm}
\end{center}
\end{table}

\subsection{Pseudo Features Analysis}
\vspace{-0.2cm}
 \begin{figure}
    \centering
      \vspace*{-0.2cm}
\includegraphics[width=0.5\textwidth]{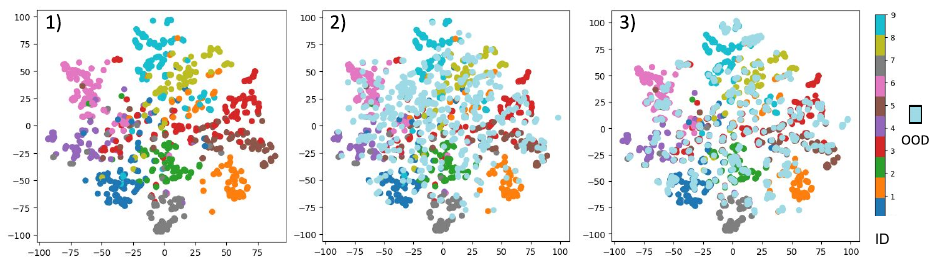}
     \vspace*{-0.7cm}
    \caption{  \footnotesize t-SNE 2D projection of the encoder features of 1) Cifar-10 ID samples, 2) real auxiliary OOD - DTD dataset - and, 3) pseudo OOD.
    }
    \label{fig:tsne}
    \vspace*{-0.4cm}
\end{figure}
In the previous experiments, we show that our method can leverage synthetic OOD data and improve the OOD detection performance with a simple approach to generate pseudo OOD features, a mixup of ID features of different classes.  Here we compare visually those synthesized features to the real OOD features of DTD dataset. Figure~\ref{fig:tsne} shows the t-SNE 2D projection of ID features (Cifar-10) at the encoder level and the OOD features for both real OOD (DTD) and pseudo OOD cases. We refer to supplementary for more details.
We can see that pseudo OOD examples act as perturbed ID samples, however denser at areas where different ID classes samples are overlapped. Training with those pseudo features encourages more compact ID clusters and hence stronger OOD detection capabilities.

\section{Conclusion}\label{sec:conclusion}
\vspace{-0.2cm}
Given the  success of supervised contrastive representation learning (\SupCon) in learning powerful representations and with the aim to overcome the known overconfidence problem by Softmax classifiers, we propose a new \OOD-aware training regime tailored for representations trained with SupCon.
We start with \SupCon loss~\cite{khosla2020supervised} and suggest to jointly learn classes prototypes as an alternative to the Softmax \CE loss.
The prototypes are optimized to be close to their corresponding class samples.
We regularize the training on \ID data with auxiliary or pseudo \OOD data.
We propose two losses operating on \OOD data, one at the projection head level and one at the encoder level.
The first operates on pairwise samples' similarities and pushes \OOD head features away from \ID head features and the later pushes \OOD encoder features far from the prototypes.
We perform experiments on a wide range of \OOD datasets and show a significant reduction on \FPR.
% Unlike previous work
Our approach does not rely on a large auxiliary \OOD dataset 
%and can be trained using a small amount of real/pseudo auxiliary data.
%Our work
and moves a step closer to deploying \OOD detector in practice by providing more reliable \OOD rejection. 

%%%%%%%%% REFERENCES
{\small
\bibliographystyle{ieee_fullname}
\bibliography{egbib}
}

\end{document}

\title{OOD Aware Supervised Contrastive Learning \\ [1ex] \large \normalfont Supplementary Materials}

\author{Soroush Seifi\\
{\tt\small soroush.seifi@external.toyota-europe.com}
\and 
Daniel Olmeda Reino\\
{\tt\small daniel.olmeda.reino@toyota-europe.com}
\and 
Nikolay Chumerin\\
{\tt\small nikolay.chumerin@toyota-europe.com}
\and
Rahaf Aljundi\\
{\tt\small rahaf.al.jundi@toyota-europe.com}\\
\and
Toyota Motor Europe
}
\maketitle

%%%%%%%%% BODY TEXT
\section{Introduction}
These supplementary materials serve as additional empirical evaluation supporting the main results in the paper. First we report the OOD performance of our method using a different architecture as a backbone, Section~\ref{sec:arch}. We then experiment with combining both real and fake OOD data, Section~\ref{sec:mix}. We continue our analysis of ID/OOD features visualization, Section~\ref{sec:tsne}. Section~\ref{sec:tim} explores the OOD detection performance when other datasets are deployed for the auxiliary OOD training. 
 \begin{figure}[t]
    \centering
 
\includegraphics[width=0.49\textwidth]{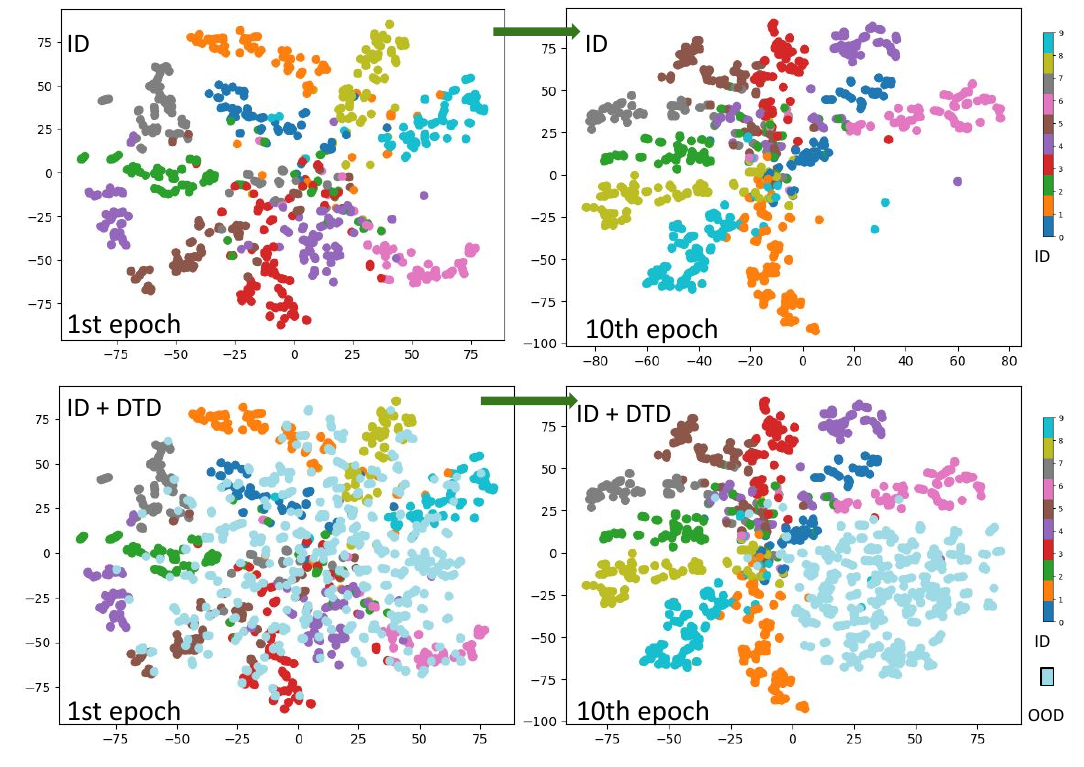}
    
    \caption{  \footnotesize t-SNE 2D projection of the encoder features of: top row, Cifar-10 ID samples; second row,  ID and real auxiliary OOD, DTD dataset. First column is for embeddings extracted at the first epoch (before OOD finetuning) and second column is after the finetuning process (10th epoch).  
    }
    \label{fig:tsne-dtd}
  
\end{figure}

 \begin{figure}[h]
    \centering
 
\includegraphics[width=0.49\textwidth]{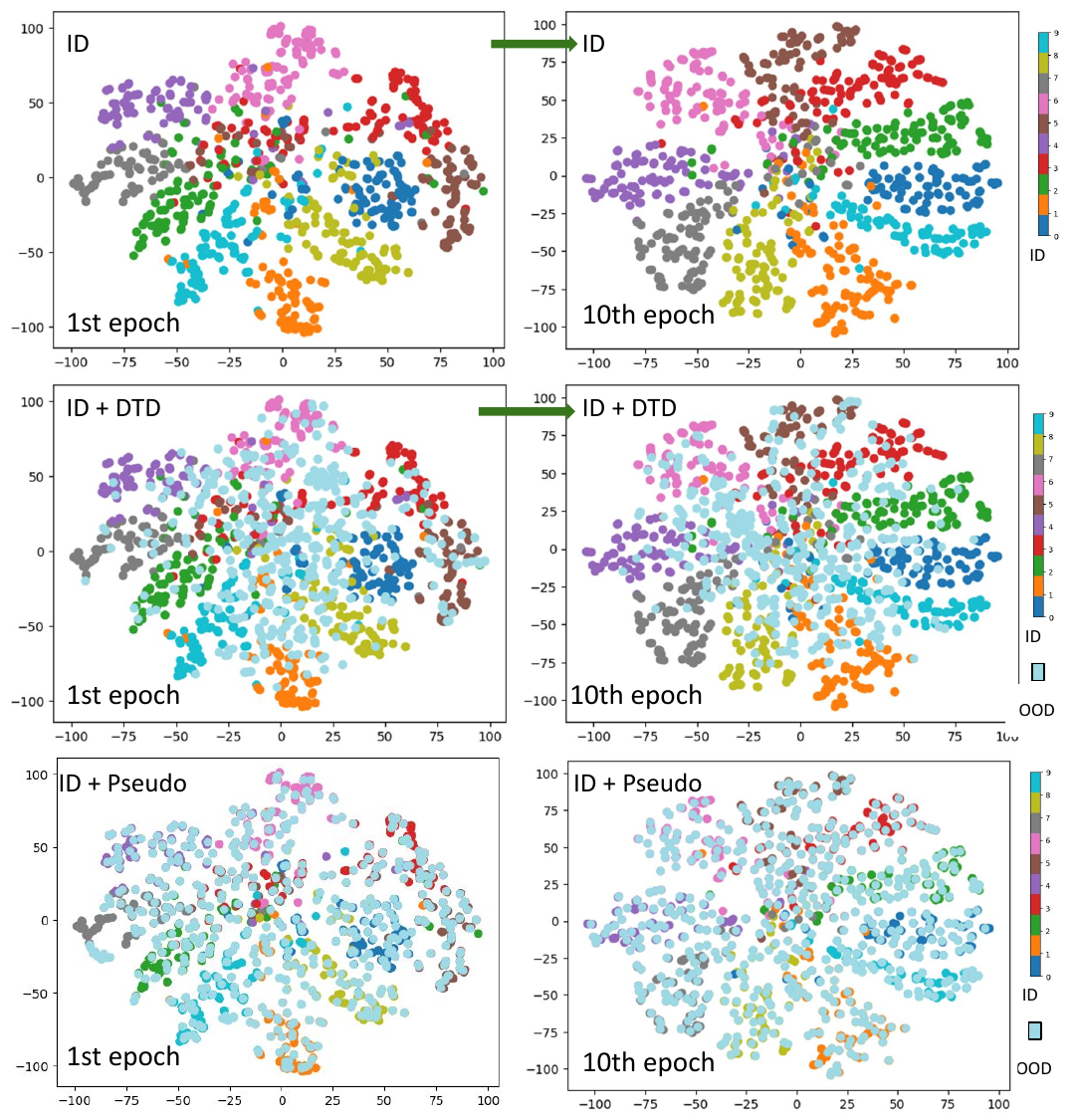 }
    
    \caption{  \footnotesize t-SNE 2D projection of the encoder features of: top row, Cifar-10 ID samples; second row, ID and real auxiliary OOD, DTD dataset; third row: ID and Pseudo OOD features.  First column is for embeddings extracted at the first epoch (before OOD finetuning) and second column is after the finetuning process (10th epoch). 
    }
    \label{fig:tsne-pseudp}
  
\end{figure}
\begin{table*}[t]
\footnotesize
\begin{center}
\resizebox{\textwidth}{!}{ 
\begin{tabular}{|l|l l l|l l l|l l l|l l l|l l l|l l l|l l l|}
\hline
Dataset/\Longunderstack{Method \\ Metrics} & \multicolumn{3}{c|}{\Longunderstack{CE \\ \\ FPR$\downarrow$ AUROC$\uparrow$  AUPR$\uparrow$ \\}} &  \multicolumn{3}{c|}{\Longunderstack{PSupCon \\ \\ FPR$\downarrow$ AUROC$\uparrow$  \AUPR$\uparrow$\\}} & \multicolumn{3}{c|}{\Longunderstack{CE + Energy \\ \\ FPR$\downarrow$ AUROC$\uparrow$  AUPR$\uparrow$ \\}} & \multicolumn{3}{c|}{\Longunderstack{PSupCon + Energy \\ \\ FPR$\downarrow$ AUROC$\uparrow$  \AUPR$\uparrow$\\}} & \multicolumn{3}{c|}{\Longunderstack{OPSupCon-R \\ \\ FPR$\downarrow$ AUROC$\uparrow$  \AUPR$\uparrow$\\}} & \multicolumn{3}{c|}{\Longunderstack{OPSupCon-P \\ \\ FPR$\downarrow$ AUROC$\uparrow$  \AUPR$\uparrow$\\}}
\\
\hline
\hline
\rowcolor{\fprcolor}
DTD & 18.17 & 95.83 & 98.79 
 & 14.70 & 97.06 & 99.30
&  \textbf{5.33} & \textbf{98.74} & \textbf{99.73}
 &  7.22 & 98.57 & 99.70
 &  10.81  &  98.13  &  99.60
 &  16.52 & 96.85 & 99.28
 \\

\rowcolor{\auroccolor}
SVHN & 2.27 & 99.44 & 99.89
 & 3.41 & 99.35 & 99.87
& 1.83 & 99.46 & 99.89
 & \textbf{0.66} & \textbf{99.81} & \textbf{99.96}
 &  2.66  &  99.42  &  99.88
 & 3.48 & 99.33 & 99.87

\\

\rowcolor{\fprcolor}
Places365 & 24.80 & 94.45 & 98.59
& 23.46 & 95.61 & 98.97
& \textbf{17.84} & 95.54 & 98.78
& 18.96 & 96.01 & 98.99
&  19.17  &  \textbf{96.17}  &  \textbf{99.09}
 &  20.14 & 96.06 & 99.06
\\

\rowcolor{\auroccolor}
LSUN-C & 2.09 & 99.37 & 99.88
& 0.24 & \textbf{99.89} & \textbf{99.98}
& 1.47 & 99.44 & 99.89
& 1.95 & 99.30 & 99.86
&  \textbf{0.21}  &  99.87  &  99.97
&  0.23  &  \textbf{99.89}  &  \textbf{99.98}

\\

\rowcolor{\fprcolor}
LSUN-R & 3.58 & 99.05 & 99.81
& \textbf{1.69} & \textbf{99.59} & \textbf{99.92}
& 4.60 & 99.03 & 99.80
& 4.96 & 98.90 & 99.78
&  2.68  &  99.40  &  99.88
&  1.80  &  99.55  &  99.91

\\

\rowcolor{\auroccolor}
iSUN & 4.19 & 99.00 & 99.80
& \textbf{1.62} & \textbf{99.59} & \textbf{99.92}
& 3.90 & 99.13 & 99.82
& 5.12 & 98.94 & 99.79
&  2.42  &  99.41  &  99.88
&  1.89  &  99.51  &  99.91
\\

\rowcolor{\fprcolor}
iNaturalist & 16.24 & 96.83 & 99.33
 & 7.98 & 98.47 & 99.69
& 9.66 & 97.73 & 99.49
 & \textbf{7.40} & \textbf{98.56} & \textbf{99.70}
&  7.94  &  98.50  &  \textbf{99.70}
&  8.94  &  98.36  &  99.67
\\

\rowcolor{\auroccolor}
CIFAR-100 & 37.77 & 92.03 & 98.03 
 & 40.61 & 93.14 & 98.52
& \textbf{31.30} & 92.87 & 98.12
 & 34.92 & 93.54 & 98.56
&  36.57  &  \textbf{93.71}  &  \textbf{98.65}
&  39.69  &  93.24  &  98.55
\\

\rowcolor{\fprcolor}
Mnist & 26.13 & 96.41 & 99.31
 & 7.16 & 98.54 & 99.72
& 19.62 & 96.87 & 99.38
 & 12.93 & 97.68 & 99.55
&  \textbf{5.78}  &  \textbf{98.82}  &  \textbf{99.77}
&  5.97  &  98.77  &  99.76
\\

\rowcolor{\auroccolor}
TIN & 28.25 & 93.56 & 98.30
 & 28.19 & 94.25 & 98.60
& 22.80 & 94.64 & 98.58
 &  \textbf{22.15} & 94.85 & 98.70
&  25.20  &  \textbf{94.96}  &  \textbf{98.77}
&  26.20 & 94.82 & 98.74
\\ 

\rowcolor{\fprcolor}
Average & 16.35 & 96.60 & 99.17
 & 12.90 & 97.55 & 99.45
& 11.83 & 97.34 & 99.35
 & 11.63 & 97.62 & 99.46
&  \textbf{11.35}  &  \textbf{97.84}  &  \textbf{99.52}
&  12.49 & 97.64 & 99.47
 \\ 
\hline
\end{tabular}}
\caption { \textbf{OOD detection performance on Cifar-10 with ResNet-50 backbone:} a) comparison of CE and PSupCon (1, 2 columns) and, b) comparison of OOD training with our method compared to energy finetuning. Our method outperforms performance of energy finetuning even with pseudo OOD.}
\label{tab:ResNet50_cifar10}
\end{center}
\vspace{-0.3cm}
\end{table*}

\begin{table*}[t]
\footnotesize
\begin{center}
\resizebox{\textwidth}{!}{ 
\begin{tabular}{|l|l l l|l l l|l l l|l l l|l l l|l l l|l l l|}
\hline
Dataset/\Longunderstack{Method \\ Metrics} & \multicolumn{3}{c|}{\Longunderstack{CE \\ \\ FPR$\downarrow$ AUROC$\uparrow$  AUPR$\uparrow$ \\}} &  \multicolumn{3}{c|}{\Longunderstack{PSupCon \\ \\ FPR$\downarrow$ AUROC$\uparrow$  \AUPR$\uparrow$\\}} & \multicolumn{3}{c|}{\Longunderstack{CE + Energy \\ \\ FPR$\downarrow$ AUROC$\uparrow$  AUPR$\uparrow$ \\}} & \multicolumn{3}{c|}{\Longunderstack{PSupCon + Energy \\ \\ FPR$\downarrow$ AUROC$\uparrow$  \AUPR$\uparrow$\\}} & \multicolumn{3}{c|}{\Longunderstack{OPSupCon-R \\ \\ FPR$\downarrow$ AUROC$\uparrow$  \AUPR$\uparrow$\\}} & \multicolumn{3}{c|}{\Longunderstack{OPSupCon-P \\ \\ FPR$\downarrow$ AUROC$\uparrow$  \AUPR$\uparrow$\\}}
\\
\hline
\hline
\rowcolor{\fprcolor}
DTD & 80.46 & 78.22 & 94.77 
 & 74.07 & 67.48 & 88.46
&  \textbf{59.08} & \textbf{87.97} & \textbf{97.30}
 &  68.14 & 85.36 & 96.77
  &  64.1 & 79.33 & 94.43
 &  65.32  &  72.88  &  90.77
 \\

\rowcolor{\auroccolor}
SVHN & 52.41 & 90.56 & 97.99
 & 85.39 & 75.30 & 94.30
& 27.71 & 95.27 & 99.01
 & \textbf{11.65} & \textbf{97.70} & \textbf{99.48}
  &  63.7 & 87.12 & 97.24
 &  92.15  &  72.65  &  93.76

\\

\rowcolor{\fprcolor}
Places365 & 81.49 & 77.14 & 94.26
& 86.33 & 71.97 & 92.78
& 77.81 & \textbf{79.87} & \textbf{95.08}
& 81.15 & 77.89 & 94.58
 &  \textbf{75.96} & 77.41 & 94.30
&  81.04  &  75.39  &  93.74
\\

\rowcolor{\auroccolor}
LSUN-C & 53.08 & 90.69 & 98.04
& 21.22 & 96.03 & 99.14
& 41.72 & 93.15 & 98.57
& 85.58 & 76.54 & 94.66
&  8.21 & 98.34 & 99.65
&  \textbf{4.67}  &  \textbf{99.01}  &  \textbf{99.79}

\\

\rowcolor{\fprcolor}
LSUN-R & 64.18 & 87.64 & 97.33
& 70.37 & 82.85 & 96.12
& 43.11 & 92.16 & 98.27
& 37.73 & 93.38 & 98.59
&  \textbf{19.43} & \textbf{96.35} & \textbf{99.21}
&  21.14  &  95.83  &  99.07

\\

\rowcolor{\auroccolor}
iSUN & 68.13 & 86.33 & 97.03
& 67.91 & 82.61 & 95.93
& 49.27 & 90.47 & 97.90
& 38.40 & 93.06 & 98.51
&  22.72 & \textbf{95.09} & \textbf{98.88}
&  \textbf{22.00}  &  94.95  &  98.80
\\

\rowcolor{\fprcolor}
iNaturalist & 85.66 & 76.57 & 94.44
 & 42.80 & 90.18 & 97.68
& 78.25 & 82.48 & 96.06
 & 68.61 & 85.25 & 96.73
 &  \textbf{34.62} & \textbf{92.30} & \textbf{98.21}
&  34.72  &  91.83  &  98.00
\\

\rowcolor{\auroccolor}
CIFAR-10& \textbf{72.06} & \textbf{82.53} & \textbf{95.87} 
 & 86.64 & 72.06 & 92.30
& 76.78 & 79.90 & 95.12
 & 89.16 & 69.95 & 91.83
&  87.34 & 69.53 & 91.22
&  88.46  &  70.19  &  91.94
\\

\rowcolor{\fprcolor}
Mnist & 94.79 & 68.66 & 92.88
 & 99.81 & 44.98 & 85.08
& 93.76 & 73.31 & 94.12
 & 95.28 & 63.57 & 91.15
&  \textbf{8.58} & \textbf{98.50} & \textbf{99.70}
&  50.05  &  90.75  &  98.06
\\

\rowcolor{\auroccolor}
TIN & 74.05 & 80.81 & 95.08
 & 77.25 & 78.64 & 94.61
& 70.95 & \textbf{82.96} & \textbf{95.61}
 &  75.48 & 80.26 & 95.04
&  \textbf{67.50} & 82.05 & 95.41
&  74.2  &  79.73  &  94.78
\\ 

\rowcolor{\fprcolor}
Average & 72.63 & 81.91 & 95.77
 & 71.18 & 76.21 & 93.64
& 61.84 & 85.75 & 96.71
 & 65.12 & 82.29 & 95.73
&  \textbf{45.21} & \textbf{87.60} & \textbf{96.82}
&  53.37  &  84.32  &  95.87
 \\ 
\hline
\end{tabular}}
\caption { \textbf{OOD detection performance on Cifar-100 with ResNet-50 backbone:} a) comparison of CE and PSupCon (1, 2 columns) and, b) comparison of OOD training with our method compared to energy finetuning. Our method outperforms performance of energy finetuning even with pseudo OOD.}
\label{tab:ResNet50_cifar100}
\end{center}
\vspace{-0.3cm}
\end{table*}

\begin{table*}[t]
\footnotesize
\begin{center}
\resizebox{\textwidth}{!}{ 
\begin{tabular}{|l|l l l|l l l|l l l|l l l|l l l|l l l|l l l|}
\hline
Dataset/\Longunderstack{Method \\ Metrics} & \multicolumn{3}{c|}{\Longunderstack{OPSupCon-R \\ MSP \\ FPR$\downarrow$ AUROC$\uparrow$  \AUPR$\uparrow$\\}} &  \multicolumn{3}{c|}{\Longunderstack{OPSupCon-R \\ Energy \\ FPR$\downarrow$ AUROC$\uparrow$  \AUPR$\uparrow$\\}} &  \multicolumn{3}{c|}{\Longunderstack{OPSupCon-R \\ Maximum logit \\ FPR$\downarrow$ AUROC$\uparrow$  \AUPR$\uparrow$\\}} & \multicolumn{3}{c|}{\Longunderstack{OPSupCon-P \\ MSP \\ FPR$\downarrow$ AUROC$\uparrow$  \AUPR$\uparrow$\\}} & \multicolumn{3}{c|}{\Longunderstack{OPSupCon-P \\ Energy \\ FPR$\downarrow$ AUROC$\uparrow$  \AUPR$\uparrow$\\}} & \multicolumn{3}{c|}{\Longunderstack{OPSupCon-P \\ Maximum logit \\ FPR$\downarrow$ AUROC$\uparrow$  \AUPR$\uparrow$\\}}
\\
\hline
\hline
\rowcolor{\fprcolor}
DTD & 7.74 & 98.58 & 99.72
 & 6.33 & 98.84 & 99.75
&  \textbf{4.95} & \textbf{99.04} & \textbf{99.80}
 &  17.60 & 97.01 & 99.39
 &  17.33 & 96.42 & 99.16
 &  16.57 & 96.69 & 99.22

 \\

\rowcolor{\auroccolor}
SVHN & 2.40 & 99.38 & 99.88
 & \textbf{0.43} & \textbf{99.87} & \textbf{99.97}
& 0.85 & 99.75 & 99.95
 & 2.71 & 99.21 & 99.84
 &  2.38 & 99.56 & 99.91
 & 5.41 & 98.46 & 99.70

\\

\rowcolor{\fprcolor}
Places365 & 21.19 & 95.82 & 98.99
& 24.40 & 95.09 & 98.78
& 21.17 & 95.63 & 98.91
& 22.75 & 95.51 & 98.94
&  27.24 & 94.96 & 98.81
 &  \textbf{14.48} & \textbf{96.76} & \textbf{99.21}
\\

\rowcolor{\auroccolor}
LSUN-C & 2.87 & 99.18 & 99.84
& 1.65 & 99.58 & \textbf{99.92}
& \textbf{1.33} & \textbf{99.60} & \textbf{99.92}
& 4.19 & 98.89 & 99.79
&  2.27 & 99.47 & 99.89
&  2.39 & 99.34 & 99.87

\\

\rowcolor{\fprcolor}
LSUN-R & 8.85 & 98.35 & 99.68
& 9.92 & 98.13 & 99.63
& 9.52 & 98.16 & 99.64
& 9.34 & 98.19 & 99.64
&  7.93 & 98.48 & 99.70
&  \textbf{6.62} & \textbf{98.57} & \textbf{99.72}

\\

\rowcolor{\auroccolor}
iSUN & 8.49 & 98.40 & 99.68
& \textbf{6.91} & 98.58 & 99.72
& 7.71 & 98.40 & 99.69
& 10.81 & 98.01 & 99.61
&  7.03 & \textbf{98.65} & \textbf{99.73}
&  7.24 & 98.52 & 99.70
\\

\rowcolor{\fprcolor}
iNaturalist & 15.45 & 97.36 & 99.48
 & \textbf{9.06} & \textbf{98.38} & \textbf{99.68}
& 9.87 & 98.11 & 99.63
 & 20.34 & 96.58 & 99.32
&  10.91 & 98.13 & 99.62
&  12.48 & 97.70 & 99.53
\\

\rowcolor{\auroccolor}
CIFAR-100 & \textbf{33.88} & \textbf{93.77} & \textbf{98.60}
 & 40.79 & 92.06 & 98.12
& 36.04 & 93.15 & 98.41
 & 36.08 & 93.39 & 98.56
&  47.67 & 91.06 & 97.97
&  36.42 & 93.25 & 98.51
\\

\rowcolor{\fprcolor}
Mnist & 13.20 & \textbf{97.87} & 99.58
 & 0.75 & 99.78 & \textbf{99.96}
& 2.79 & 99.42 & 99.89
 & 13.73 & 97.74 & 99.56
&  \textbf{0.55} & 99.70 & 99.94
&  8.10 & 98.55 & 99.72

\\

\rowcolor{\auroccolor}
TIN & 26.91 & 94.17 & 98.56
 & 30.29 & 93.23 & 98.25
& 25.83 & 94.39 & 98.61
 &  28.38 & 94.03 & 98.56
&  33.22 & 93.17 & 98.29
&  \textbf{25.55} & \textbf{94.61} & \textbf{98.64}
\\ 

\rowcolor{\fprcolor}
Average & 14.09 & 97.29 & 99.40
 & 13.05 & 97.35 & 99.38
& \textbf{12.01} & \textbf{97.56} & \textbf{99.44}
 & 16.59 & 96.86 & 99.32
& 15.65 & 96.96 & 99.30
& \textbf{13.52} & \textbf{97.24} & \textbf{99.38}
 \\ 
\hline
\end{tabular}}
\caption {\textbf{Ablation on different scoring functions.} Maximum logit score achieves the best average results.}
\label{tab:ResNet18_cifar10_scoring}
\end{center}
\vspace{-0.3cm}
\end{table*}
\section{Another Backbone}\label{sec:arch}
In order to have a fair comparison with previous work, in the main paper we show results with a ResNet18 backbone. Here we investigate the effect of changing the backbone to a larger network, namely ResNet50. 

Similar to the main experiments in the main paper, models are trained for 500 epochs. We notice that with ResNet50 our method requires less number of epochs for finetuning.  For \oursreal and \oursfake , we finetune PSupCon for 25 and 10 epochs on DTD \cite{cimpoi2014describing} and pseudo OOD features respectively. We observe that the performance improves over PSupCon from the very first epochs of finetuning.

Tables \ref{tab:ResNet50_cifar10} and \ref{tab:ResNet50_cifar100} follow the same trend as the results reported in the main paper for different models. This suggests that our proposed method is robust to changes in the feature extractor. Especially, on the more challenging CIFAR-100 \cite{krizhevsky2009learning} dataset, our method improves over Energy finetuning \cite{liu2020energy} with a large margin, for both auxiliary (\oursreal) and pseudo (\oursfake) OOD training: 7\% reduction in FPR and  16\% reduction in FPR respectively.

\section{Mixed OPSupCon}\label{sec:mix}

\begin{table}
\scriptsize
\centering
\setlength{\extrarowheight}{0pt}
\addtolength{\extrarowheight}{\aboverulesep}
\addtolength{\extrarowheight}{\belowrulesep}
\setlength{\aboverulesep}{0pt}
\setlength{\belowrulesep}{0pt}
\begin{tabular}{|p{1.1cm}|p{0.6cm}|p{\metricsize}p{\metricsize}p{\metricsize}p{\metricsize}p{\metricsize}p{\metricsize}p{\metricsize}|} 
\toprule
Method                                           & Metric      & \begin{sideways}DTD\end{sideways}        & \multicolumn{1}{|l|}{\begin{sideways}SVHN\end{sideways}} & \multicolumn{1}{l|}{\begin{sideways}Places365\end{sideways}} & \multicolumn{1}{l|}{\begin{sideways}CIFAR-100\end{sideways}} & \multicolumn{1}{l|}{\begin{sideways}MNIST\end{sideways}} & \multicolumn{1}{l|}{\begin{sideways}TIN\end{sideways}} & \begin{sideways}Average\end{sideways}     \\ 
\hline
\hline
\multirow{3}{*}{\Longunderstack{\ours\\R}} & {\cellcolor{\fprcolor}}{\tiny FPR$\downarrow$}   & {\cellcolor{\fprcolor}} 8.27 & {\cellcolor{\fprcolor}} 3.27                 & {\cellcolor{\fprcolor}} 21.98                    & {\cellcolor{\fprcolor}} 43.70                      & {\cellcolor{\fprcolor}} 6.46                  & {\cellcolor{\fprcolor}} 33.12                & {\cellcolor{\fprcolor}} 19.46  \\
                                                 & {\cellcolor{\auroccolor}}{\tiny AUROC$\uparrow$} & {\cellcolor{\auroccolor}}  98.48  & {\cellcolor{\auroccolor}} 99.26                & {\cellcolor{\auroccolor}} 95.37                     & {\cellcolor{\auroccolor}}  91.20                      & {\cellcolor{\auroccolor}} 98.58                  & {\cellcolor{\auroccolor}} 93.40                & {\cellcolor{\auroccolor}} 96.04   \\
                                                 & {\cellcolor{\auprcolor}}{\tiny AUPR$\uparrow$}  & {\cellcolor{\auprcolor}} \textbf{99.68}  & {\cellcolor{\auprcolor}} 99.85               & {\cellcolor{\auprcolor}} 98.83                     & {\cellcolor{\auprcolor}} 97.87                      & {\cellcolor{\auprcolor}} 99.72                  & {\cellcolor{\auprcolor}} 98.36                & {\cellcolor{\auprcolor}} 99.05   \\ 
\hline
\multirow{3}{*}{\Longunderstack{\ours\\P}} & {\cellcolor{\fprcolor}}{\tiny FPR$\downarrow$}   & {\cellcolor{\fprcolor}} 18.65  & {\cellcolor{\fprcolor}} 4.88                & {\cellcolor{\fprcolor}} 25.02                      & {\cellcolor{\fprcolor}} 46.43                      & {\cellcolor{\fprcolor}} \textbf{4.48}                 & {\cellcolor{\fprcolor}} 34.23               & {\cellcolor{\fprcolor}} 22.28    \\
                                                 & {\cellcolor{\auroccolor}}{\tiny AUROC$\uparrow$} & {\cellcolor{\auroccolor}}  96.11   & {\cellcolor{\auroccolor}} 99.0                & {\cellcolor{\auroccolor}}  95.00                     & {\cellcolor{\auroccolor}} 90.48                      & {\cellcolor{\auroccolor}} \textbf{98.97}                & {\cellcolor{\auroccolor}} 93.16                & {\cellcolor{\auroccolor}} 95.45  \\
                                                 & {\cellcolor{\auprcolor}}{\tiny AUPR$\uparrow$}  & {\cellcolor{\auprcolor}} 99.07  & {\cellcolor{\auprcolor}} 99.80                & {\cellcolor{\auprcolor}}98.79                    & {\cellcolor{\auprcolor}} 97.78                     & {\cellcolor{\auprcolor}} \textbf{99.80}                 & {\cellcolor{\auprcolor}} 98.30                & {\cellcolor{\auprcolor}}98.92  \\ 
\hline
\multirow{3}{*}{\Longunderstack{\ours\\M}} & {\cellcolor{\fprcolor}}{\tiny FPR$\downarrow$}   & {\cellcolor{\fprcolor}} \textbf{8.22}  & {\cellcolor{\fprcolor}} \textbf{2.51}               & {\cellcolor{\fprcolor}} \textbf{20.34}                    & {\cellcolor{\fprcolor}} \textbf{43.21}                     & {\cellcolor{\fprcolor}} 4.95              & {\cellcolor{\fprcolor}} \textbf{31.48}              & {\cellcolor{\fprcolor}} \textbf{18.45}    \\
                                                 & {\cellcolor{\auroccolor}}{\tiny AUROC$\uparrow$} & {\cellcolor{\auroccolor}} \textbf{98.49}  & {\cellcolor{\auroccolor}}\textbf{99.40}                & {\cellcolor{\auroccolor}}  \textbf{95.65}                     & {\cellcolor{\auroccolor}} \textbf{91.30}                     & {\cellcolor{\auroccolor}} 98.92 & {\cellcolor{\auroccolor}} \textbf{93.58}                & {\cellcolor{\auroccolor}} \textbf{96.22}  \\
                                                 & {\cellcolor{\auprcolor}}{\tiny AUPR$\uparrow$}  & {\cellcolor{\auprcolor}} \textbf{99.68}  & {\cellcolor{\auprcolor}}  \textbf{99.88}              & {\cellcolor{\auprcolor}} \textbf{98.88}                    & {\cellcolor{\auprcolor}} \textbf{97.89}                      & {\cellcolor{\auprcolor}}99.78               & {\cellcolor{\auprcolor}} \textbf{98.38}                & {\cellcolor{\auprcolor}} \textbf{99.08}   \\ 

%24.74 AUC 94.31 AUPIN 93.43
\bottomrule
\end{tabular}
\caption {\textbf{Comparison of our method's variants on CIFAR-10 dataset.} OpSupCon-M represents using both real auxiliary OOD (DTD) data and our pseudo OOD features when generating OOD training samples.  }
\label{tab:mixed_cifar10}
\end{table}
In the main paper, we show that in case OOD data cannot be gathered or synthetically generated, pseudo OOD data can be generated using a simple mixup of the ID features of different classes. Here, we further evaluate the performance of our method when generating OOD training data by combining real OOD features (Textures dataset, DTD) with pseudo OOD features. We use our complete loss to finetune PSupCon with such data and name this model as OPSupCon-M (as for Mixed-OOD).  Table~\ref{tab:mixed_cifar10}  reports the performance of our method when leveraging different types of OOD data.
Combining real auxiliary OOD with pseudo OOD adds a further boost and robustness to the OOD detection performance. 

% \section{Finetuning vs End-to-End Training}\label{sec:training}
% Here we report the accuracy of the OPSupCon model when trained from scratch with the loss defined in equation 6 in the main paper. While most of the previous work finetune their model for the OOD detection task after pre-training a classifier on the ID dataset, our model can be trained simultaneously for both classification and OOD detection tasks. Table \ref{ftvssc} suggest that the performance would not degrade in this case.

% \input{tables/supp-tables/FTvsScratch}

% \section{Other OOD datasets for training}\label{sec:data}

% In this section we provide ablation on using other datasets as well as Tinyimages for the task of real OOD training. Similar to the settings in the main paper, our experiments are done on CIFAR-10 dataset with a ResNet18 backbone, table \ref{tab:other_ood}. This table suggest that the OOD detection performance is not dependant on the size of the dataset with Tinyimages (TIM) being the largest dataset and performing worse on most of the metrics compared to the other datasets. Besides, the performance gap between the experiments on average may be negligible while some perform better on individual datasets (i.e FPR on DTD, iSUN and iNaturalist). However, in order to stay consistent with previous work that train with real-OOD data, we report our results in the main paper using TIM as the OOD dataset.

% \input{tables/supp-tables/other-ood}
\section{Encoder Features Analysis}\label{sec:tsne}
In the main paper, we analyze the features of ID, auxiliary and pseudo OOD  samples with a t-SNE 2D projection. However, we only compared ID and OOD features before starting the finetuning process with our method. Here, we analyze those features \textit{after} finetuning with our method. We consider a ResNet18 model trained for 100 epochs on Cifar-10 dataset. We train our \oursreal and \oursfake for 10 epochs.

Figure~\ref{fig:tsne-dtd} visualizes the 2D projections of ID features and auxiliary OOD features from DTD datasets at the beginning and at the end of the finetuning process for~\oursreal. We see that features from the OOD dataset are initially projected quite close to the ID features of Cifar-10 dataset which makes the OOD detection difficult. After the model is finetuned, the OOD features from DTD dataset are projected into a cluster clearly separate from the ID features. This results in a significant improvements on the OOD detection performance. 

Figure~\ref{fig:tsne-pseudp} visualizes the t-SNE 2D projection of ID features, real OOD features from DTD and the generated pseudo OOD features both at the beginning and at the last epoch of the training for \oursfake. We can draw the following observations on the results of fientuning with \oursfake : 
\begin{itemize}
    \item The ID features clusters are more compact with a lesser of an overlap (middle of the plot).
    \item The OOD features of DTD are pushed further away from the dense areas of ID clusters in spite of not being trained explicitly on those features.
    \item The pseudo generated features get more difficult to distinguish from ID data as we proceed with the training.
\end{itemize}
Indeed the pseudo generated features act as a regularization to the ID features pushing samples of the same class to be closer together and further from other classes samples. As pseudo OOD samples are generated on the fly, while ID clusters get more compact, it gets more difficult for the model to distinguish them from the actual ID data. This is due to the fact that pseudo OOD features become more and more similar to those of ID dataset as the training goes on. Consequently, we observed that training \oursfake for a few epochs is enough to achieve a good OOD performance while training for a large number of epochs might have a negative effect instead.

\section{Effect of the choice of Auxiliary \OOD Data  } \label{sec:tim}

\begin{table}
\scriptsize
\centering
\setlength{\extrarowheight}{0pt}
\addtolength{\extrarowheight}{\aboverulesep}
\addtolength{\extrarowheight}{\belowrulesep}
\setlength{\aboverulesep}{0pt}
\setlength{\belowrulesep}{0pt}
\begin{tabular}{|p{1.1cm}|p{0.6cm}|p{\metricsize}p{\metricsize}p{\metricsize}p{\metricsize}p{\metricsize}p{\metricsize}p{\metricsize}|} 
\toprule
Method & Metric & \begin{sideways}DTD\end{sideways} & \multicolumn{1}{|l|}{\begin{sideways}SVHN\end{sideways}} & \multicolumn{1}{l|}{\begin{sideways}Places365\end{sideways}} & \multicolumn{1}{l|}{\begin{sideways}CIFAR-100\end{sideways}} & \multicolumn{1}{l|}{\begin{sideways}MNIST\end{sideways}} & \multicolumn{1}{l|}{\begin{sideways}TIN\end{sideways}} & \begin{sideways}Average\end{sideways}     \\ 
\hline
\hline
\multirow{3}{*}{\Longunderstack{PSupCon}}& {\cellcolor{\fprcolor}}{\tiny FPR$\downarrow$}   & {\cellcolor{\fprcolor}} 20.44 & {\cellcolor{\fprcolor}} 5.32               & {\cellcolor{\fprcolor}} 26.38                  & {\cellcolor{\fprcolor}} 47.62                     & {\cellcolor{\fprcolor}} 5.34               & {\cellcolor{\fprcolor}} 35.60               & {\cellcolor{\fprcolor}} 23.45  \\
                                                 & {\cellcolor{\auroccolor}}{\tiny AUROC$\uparrow$} & {\cellcolor{\auroccolor}}  96.04 & {\cellcolor{\auroccolor}}  98.99                 & {\cellcolor{\auroccolor}} 94.85                    & {\cellcolor{\auroccolor}}  90.47                     & {\cellcolor{\auroccolor}}  98.81                  & {\cellcolor{\auroccolor}} 92.92               & {\cellcolor{\auroccolor}} 95.34  \\
                                                 & {\cellcolor{\auprcolor}}{\tiny AUPR$\uparrow$}  & {\cellcolor{\auprcolor}} 99.09 & {\cellcolor{\auprcolor}} 99.80               & {\cellcolor{\auprcolor}} 98.75                    & {\cellcolor{\auprcolor}} 97.27                   & {\cellcolor{\auprcolor}} 94.81                 & {\cellcolor{\auprcolor}} 98.00              & {\cellcolor{\auprcolor}} 97.95 \\ 

\hline
\hline
\multirow{3}{*}{\Longunderstack{DTD}}& {\cellcolor{\fprcolor}}{\tiny FPR$\downarrow$}   & {\cellcolor{\fprcolor}} \textbf{8.27} & {\cellcolor{\fprcolor}} 3.27               & {\cellcolor{\fprcolor}} \textbf{21.98}                   & {\cellcolor{\fprcolor}} \textbf{43.70}                     & {\cellcolor{\fprcolor}} 6.46                & {\cellcolor{\fprcolor}} \textbf{33.12}               & {\cellcolor{\fprcolor}} \textbf{19.46}  \\
                                                 & {\cellcolor{\auroccolor}}{\tiny AUROC$\uparrow$} & {\cellcolor{\auroccolor}}  \textbf{98.48} & {\cellcolor{\auroccolor}}  99.26                 & {\cellcolor{\auroccolor}} \textbf{95.37}                     & {\cellcolor{\auroccolor}}  \textbf{91.20}                     & {\cellcolor{\auroccolor}}  98.58                  & {\cellcolor{\auroccolor}} 93.40               & {\cellcolor{\auroccolor}} \textbf{96.04}  \\
                                                 & {\cellcolor{\auprcolor}}{\tiny AUPR$\uparrow$}  & {\cellcolor{\auprcolor}} \textbf{99.68} & {\cellcolor{\auprcolor}} 99.85               & {\cellcolor{\auprcolor}} 98.83                    & {\cellcolor{\auprcolor}} 97.87                    & {\cellcolor{\auprcolor}} 99.72                 & {\cellcolor{\auprcolor}} 98.36              & {\cellcolor{\auprcolor}} \textbf{99.21}  \\ 

\hline
\multirow{3}{*}{\Longunderstack{TIN}} & {\cellcolor{\fprcolor}}{\tiny FPR$\downarrow$} & 
{\cellcolor{\fprcolor}} 19.81 & {\cellcolor{\fprcolor}} \textbf{2.53} & {\cellcolor{\fprcolor}} 25.82 & {\cellcolor{\fprcolor}} 47.19 & {\cellcolor{\fprcolor}} \textbf{1.93} & {\cellcolor{\fprcolor}} 33.53 & {\cellcolor{\fprcolor}} 21.80 \\

& {\cellcolor{\fprcolor}}{\tiny AUROC$\uparrow$} & 
{\cellcolor{\auroccolor}} 96.66 & {\cellcolor{\auroccolor}} \textbf{99.43} & {\cellcolor{\auroccolor}} 95.11 & {\cellcolor{\auroccolor}} 91.14 & {\cellcolor{\auroccolor}} \textbf{99.55} & {\cellcolor{\auroccolor}}  \textbf{94.03}   & {\cellcolor{\auroccolor}} 95.98  \\

& {\cellcolor{\auroccolor}}{\tiny AUPR$\uparrow$}  &
{\cellcolor{\auprcolor}} 99.30 & {\cellcolor{\auprcolor}} \textbf{99.89} & {\cellcolor{\auprcolor}} \textbf{98.86} & {\cellcolor{\auprcolor}} \textbf{97.99} & {\cellcolor{\auprcolor}} \textbf{99.91} & {\cellcolor{\auprcolor}} \textbf{98.67} & {\cellcolor{\auprcolor}} 99.10  \\ 

\hline
\end{tabular}
\vspace{0.1cm}
\caption {\textbf{OOD detection performance when different auxiliary OOD datasets are employed for training: ID dataset is CIFAR-10.} \FPR$\downarrow$, \AUROC$\uparrow$ and \AUPR$\uparrow$.  }%The OOD detection performance is better for the specific trained dataset, our OOD regularized training doesn't strongly depend on the available auxiliary OOD data and improves on other datasets as well.
\label{tab:timsun}
\vspace{-0.3cm}
\end{table}
In the main paper, we consider DTD (textures) dataset for training \oursreal. This section investigates the effect of selecting another \OOD dataset on the performance. 

Here we test \oursreal with TinyImagenet (TIN) \cite{le2015tiny} dataset which combines 200 different object categories and is similar in nature to CIFAR datasets.  Table~\ref{tab:timsun} summarises the OOD detection performance of our model trained on different \OOD datasets for CIFAR-10 as the \ID task.
% Depending on the \OOD dataset being evaluated, \OOD training with one dataset can improve over the other.
% This is probably related to what extent the deployed auxiliary \OOD data is similar to the test \OOD data.
% For example, training with TIN can improve the \OOD detection on CIFAR-100 compared to DTD based training while training on DTD provides perfect \OOD detection of MNIST dataset.

We observe that training with TIN dataset improves the OOD detection performance over plain PSupCon on all datasets.
However, training with DTD results in a  better OOD detection performance as this is a generic dataset and does not represent specific objects. It is worth noting that this is a beneficial property as a similar dataset to DTD can be easily generated synthetically . 
% However, on average the OOD detection performance is quite close for both OOD datasets in spite of the large difference in scale. This possibley suggests that a large collection of OOD data is not essential. 
\section{Choice of the scoring function  } \label{sec:scoring}

In the main paper, we consider Maximum Logit~\cite{hendrycks2019scaling} as our scoring function. This section investigates the effect of selecting two other commonly used scoring functions namely Maximum Softmax Probability \cite{hendrycks2016baseline} and (Sum) Energy \cite{liu2020energy} score  for detecting OOD examples. 

We observe that on average Maximum Logit score achieves the best OOD detection performance for both \oursreal and \oursfake models. This is due to the fact that the maximum logit measures the distance to the class prototypes which is the metric being optimized during OOD training in our method. 

\section{Comparison with SSD \cite{sehwag2021ssd}}

We compare our method against various state-of-the-art works in tables 3 and 4 of the main paper and show \oursreal performs the best compared to methods from different lines of literature.

We notice that \oursreal achieves an overall lower performance on FPR and AUROC metrics for the CIFAR-100 dataset comapred to the self-supervised method proposed in \cite{sehwag2021ssd}. This is mainly due to the performance gap on the SVHN dataset. Our method achieves better results on the majority of the other datasets.

In this section, we extensively compare our method to SSD with the settings defined in section 4.1 of the main paper. This is the optimal default setting for both \oursfake and SSD \cite{sehwag2021ssd}. Besides, we evaluate the performance on a larger number of datasets here. 

As shown in tables \ref{tab:ResNet18_cifar10_SSD} and \ref{tab:ResNet18_cifar100_SSD}, \oursfake outperforms SSD on the large majority of the datasets achieving a much better average on all metrics. Therefore, we confirm that the slightly better overall performance of SSD on table 4 of the main paper is justified by the smaller number of evaluated datasets and SSD's superior performance on the SVHN dataset.

%\section{Comparison with SSD} \label{sec:SSD}

\begin{table}[t]
\footnotesize
\begin{center}
\resizebox{0.5\textwidth}{!}{ 
\begin{tabular}{|l|l l l|l l l|l l l|l l l|l l l|l l l|l l l|}
\hline
Dataset/\Longunderstack{Method \\ Metrics} & \multicolumn{3}{c|}{\Longunderstack{OPSupCon-R \\ \\ FPR$\downarrow$ AUROC$\uparrow$  \AUPR$\uparrow$\\}} &  \multicolumn{3}{c|}{\Longunderstack{OPSupCon-p \\  \\ FPR$\downarrow$ AUROC$\uparrow$  \AUPR$\uparrow$\\}} & \multicolumn{3}{c|}{\Longunderstack{SSD \\ SupCon \\ FPR$\downarrow$ AUROC$\uparrow$  \AUPR$\uparrow$\\}}
\\
\hline
\hline
\rowcolor{\fprcolor}
DTD & \textbf{4.95} & \textbf{99.04} & \textbf{99.80}
 &  16.57 & 96.69 & 99.22
   & 10.01 & 98.29 & 97.00

 \\

\rowcolor{\auroccolor}
SVHN & 0.85 & 99.75 & 99.95
 & 5.41 & 98.46 & 99.70
    & \textbf{0.41} & \textbf{99.89} & \textbf{99.96}
\\ 

\rowcolor{\fprcolor}
Places365 & 21.17 & 95.63 & 98.91
& \textbf{14.48} & \textbf{96.76} & 99.21
   & 28.62 & 94.46 & \textbf{99.77} 
\\ 

\rowcolor{\auroccolor}
LSUN-C & \textbf{1.33} & \textbf{99.60} & \textbf{99.92}
& 2.39 & 99.34 & 99.87
   & 6.76 & 98.57 & 98.21
\\ 

\rowcolor{\fprcolor}
LSUN-R & 9.52 & 98.16 & 99.64
& \textbf{6.62} & \textbf{98.57} & \textbf{99.72}
   & 68.61 & 90.44 & 84.28
\\ 

\rowcolor{\auroccolor}
iSUN & 7.71 & 98.40 & 99.69
& \textbf{7.24} & \textbf{98.52} & \textbf{99.70}
   & 69.98 & 89.51 & 82.24
\\ 

\rowcolor{\fprcolor}
iNaturalist & \textbf{9.87} & \textbf{98.11} & \textbf{99.63}
& 12.48 & 97.70 & 99.53
   & 37.18 & 94.63 & 92.86
\\ 

\rowcolor{\auroccolor}
CIFAR-100 & \textbf{36.04} & 93.15 & 98.41
&  36.42 & \textbf{93.25} & \textbf{98.51}
   & 43.03 & 91.60 & 90.70
\\ 

\rowcolor{\fprcolor}
Mnist & \textbf{2.79} & \textbf{99.42} & \textbf{99.89}
& 8.10 & 98.55 & 99.72
   & 13.11 & 98.04 & 97.72
\\ 

\rowcolor{\auroccolor}
TIN & 25.83 & 94.39 & 98.61
& \textbf{25.55} & \textbf{94.61} & \textbf{98.64}
   & 34.62 & 92.62 & 92.20
\\

\rowcolor{\fprcolor}
Average & \textbf{12.01} & \textbf{97.56} & \textbf{99.44}
& 13.52 & 97.24 & 99.38
   & 31.23 &  94.80 & 93.49
 \\ 
\hline
\end{tabular}
}
\caption {SSD Comparison ResNet-18 CIFAR-10.}
\label{tab:ResNet18_cifar10_SSD}
\end{center}
\vspace{-0.3cm}
\end{table}
\begin{table}[t]
\footnotesize
\begin{center}
\resizebox{0.5\textwidth}{!}{ 
\begin{tabular}{|l|l l l|l l l|l l l|l l l|l l l|l l l|l l l|}
\hline
Dataset/\Longunderstack{Method \\ Metrics} & \multicolumn{3}{c|}{\Longunderstack{OPSupCon-R \\ \\ FPR$\downarrow$ AUROC$\uparrow$  \AUPR$\uparrow$\\}} &  \multicolumn{3}{c|}{\Longunderstack{OPSupCon-p \\  \\ FPR$\downarrow$ AUROC$\uparrow$  \AUPR$\uparrow$\\}} & \multicolumn{3}{c|}{\Longunderstack{SSD \\ SupCon \\ FPR$\downarrow$ AUROC$\uparrow$  \AUPR$\uparrow$\\}}
\\
\hline
\hline
\rowcolor{\fprcolor}
DTD & 51.22 & 88.44 & \textbf{97.28}
& 54.23 & 84.77 & 95.89
  & \textbf{50.19} & \textbf{90.79} & 83.24

 \\

\rowcolor{\auroccolor}
SVHN & 44.26 & 92.39 & 98.39
& 49.49 & 90.89 & 98.04
  & \textbf{11.77} & \textbf{97.87} & \textbf{99.17}
\\ 

\rowcolor{\fprcolor}
Places365 & 74.52 & 79.30 & 94.79
& \textbf{74.45} & \textbf{79.71} & \textbf{94.95}
  & 79.30 & 76.64 & 98.86
 \\ 

 \rowcolor{\auroccolor}
LSUN-C & 20.38 & 96.48 & 99.27
& \textbf{18.10} & \textbf{96.71} & \textbf{99.30}
  & 42.34 & 93.53 & 91.62
 \\ 

 \rowcolor{\fprcolor}
LSUN-R & 38.54 & \textbf{93.01} & \textbf{98.49}
& \textbf{37.85} & 92.78 & 98.43
  & 84.85 & 81.57 & 74.13
 \\ 
 
 \rowcolor{\auroccolor}
iSUN & 46.45 & \textbf{91.33} & \textbf{98.13}
& \textbf{46.38} & 90.82 & 97.97
  & 86.46 & 80.52 & 70.54
 \\ 

  \rowcolor{\fprcolor}
iNaturalist & 47.71 & 89.87 & 97.63
& \textbf{45.38} & \textbf{89.97} & \textbf{97.64}
  & 73.87 & 82.44 & 78.91
 \\

\rowcolor{\auroccolor}
CIFAR-10 & 84.74 & 71.01 & 91.50
& \textbf{84.08} & \textbf{73.11} & \textbf{92.73}
  & 87.24 & 69.82 & 66.21
\\ 

 \rowcolor{\auroccolor}
Mnist & 33.89 & \textbf{94.38} & \textbf{98.83}
& \textbf{33.78} & 94.37 & \textbf{98.83}
  & 55.20 & 89.09 & 87.09
 \\ 

  \rowcolor{\auroccolor}
TIN & \textbf{68.0} & \textbf{82.67} & \textbf{95.52}
& 69.23 & 82.12 & 95.44
  & 74.91 & 80.19 & 77.33
 \\

\rowcolor{\fprcolor}
Average & \textbf{50.97} & \textbf{87.89} & \textbf{96.98}
& 51.29 & 87.53 & 96.92
  & 63.71 & 84.24 & 82.71
 \\ 
\hline
\end{tabular}}
\caption {SSD Comparison ResNet-18 CIFAR-100.}
\label{tab:ResNet18_cifar100_SSD}
\end{center}
\vspace{-0.3cm}
\end{table}

{\small
\bibliographystyle{ieee_fullname}
\bibliography{egbib}
}